\documentclass[lettersize,journal]{IEEEtran}
\usepackage{amsmath,amsfonts}
\usepackage{algorithmic}
\usepackage{algorithm}
\usepackage{array}
\usepackage{bbm}
\usepackage{booktabs} 
\usepackage{caption}
\usepackage{cite}
\usepackage{color}
\usepackage{graphicx}
\usepackage{hyperref}
\usepackage{makecell}
\usepackage{multicol}
\usepackage{multirow}
\usepackage[caption=false,font=normalsize,labelfont=sf,textfont=sf]{subfig}
\usepackage{textcomp}
\usepackage{stfloats}
\usepackage{url}
\usepackage{verbatim}
\definecolor{Red}{cmyk}{0,1,1,0}
\definecolor{Green}{cmyk}{1,0,1,0}
\definecolor{Cyan}{cmyk}{1,0,0,0}
\definecolor{Purple}{cmyk}{0.45,0.86,0,0}
\definecolor{Sky}{cmyk}{1.0,0.8,0,0}
\definecolor{Rosolic}{cmyk}{0.00,1.00,0.50,0}
\definecolor{Blue}{cmyk}{1.00,1.00,0.00,0}
\definecolor{Orange}{cmyk}{0,0.52,0.80,0}
\definecolor{Black}{cmyk}{1,0,0,1}

\begin{document}

\title{A Text-to-3D Framework for Joint Generation of CG-Ready Humans and Compatible Garments}

\author{Zhiyao~Sun,
        Yu-Hui Wen,
        Ho-Jui Fang,
        Sheng~Ye,
        Matthieu Lin,
        Tian Lv,
        and~Yong-Jin~Liu
        \vspace{5pt} \\
        {\tt\small Project Page: \url{https://human-tailor.github.io}}
\thanks{Corresponding authors: Yong-Jin Liu and Yu-Hui Wen.}}



\maketitle

\begin{abstract}
Creating detailed 3D human avatars with fitted garments traditionally requires specialized expertise and labor-intensive workflows. While recent advances in generative AI have enabled text-to-3D human and clothing synthesis, existing methods fall short in offering accessible, integrated pipelines for generating CG-ready 3D avatars with physically compatible outfits; here we use the term CG-ready for models following a technical aesthetic common in computer graphics (CG) and adopt standard CG polygonal meshes and strands representations (rather than neural representations like NeRF and 3DGS) that can be directly integrated into conventional CG pipelines and support downstream tasks such as physical simulation. To bridge this gap, we introduce Tailor, an integrated text-to-3D framework that generates high-fidelity, customizable 3D avatars dressed in simulation-ready garments. Tailor consists of three stages. (1) Semantic Parsing: we employ a large language model to interpret textual descriptions and translate them into parameterized human avatars and semantically matched garment templates. (2) Geometry-Aware Garment Generation: we propose topology-preserving deformation with novel geometric losses to generate body-aligned garments under text control. (3) Consistent Texture Synthesis: we propose a novel multi-view diffusion process optimized for garment texturing, which 
enforces view consistency, preserves photorealistic details, and optionally supports symmetric texture generation common in garments. Through comprehensive quantitative and qualitative evaluations, we demonstrate that Tailor outperforms state-of-the-art methods in fidelity, usability, and diversity. Our code will be released for academic use.
\end{abstract}


\begin{figure*}[!h]
    \centering
    \includegraphics[width=1\linewidth]{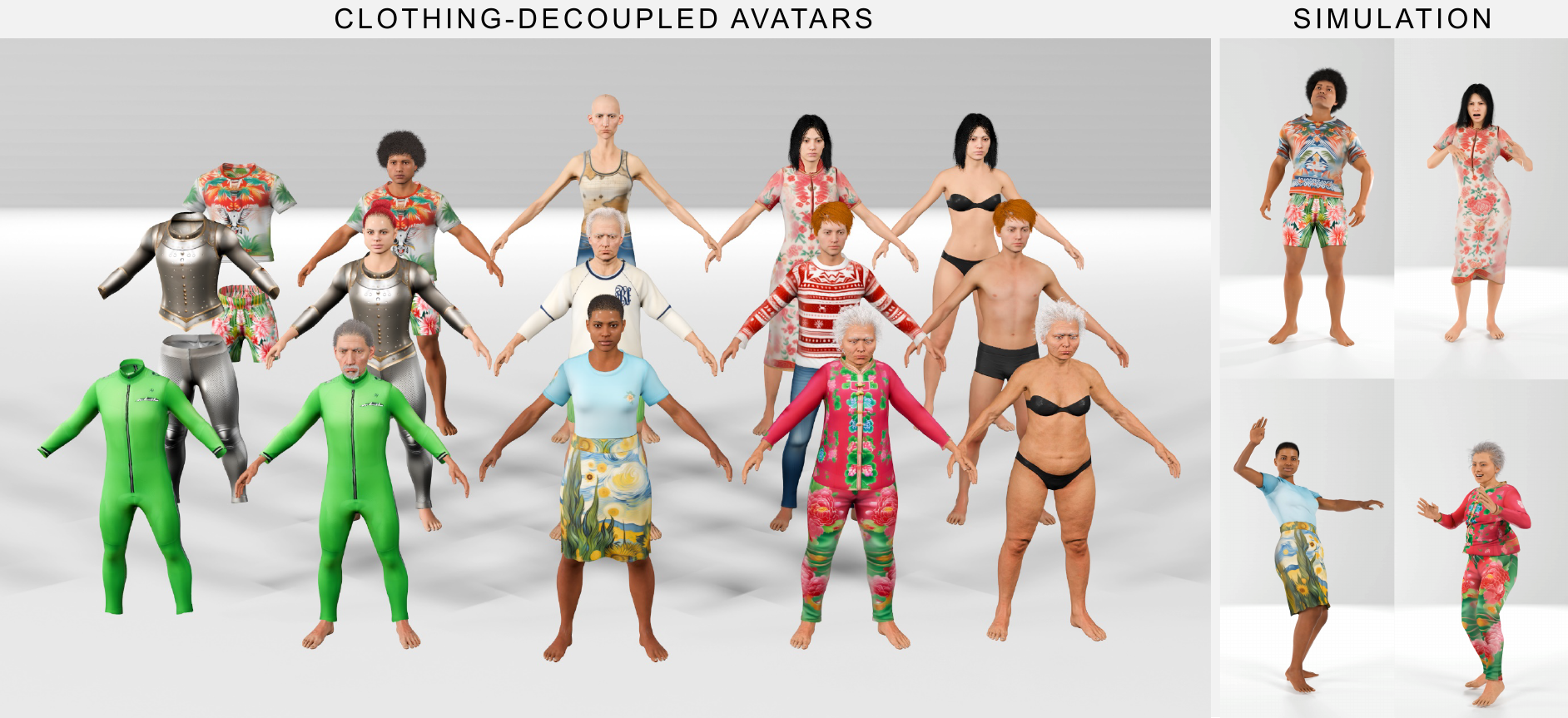}
    \caption{We introduce Tailor, an integrated text-to-3D framework that generates high-fidelity, customizable 3D humans with simulation-ready garments.}
    \label{fig:teaser}
\end{figure*}

\section{Introduction}
\label{sec:intro}

The creation of animatable 3D human avatars 
has long been an essential need in industries such as gaming, film, and virtual reality. Traditionally, this creation requires labor-intensive manual modeling by skilled artists. As the demand for such creations expands across a growing array of applications, there is an increasing need for more efficient and accessible methods of avatar production. Recent advances in generative artificial intelligence~\cite{SDS,lin2023magic3d} have led to significant progress in text-to-3D  human and garment 
generation, offering the potential to automate this process.

However, existing approaches still face several critical challenges that impede their practical adoption and overall effectiveness. For 3D avatar generation, many existing methods rely on Score Distillation Sampling (SDS)~\cite{DBLP:conf/mm/HuangYLY023, SO-SMPL}. 
While SDS enables text-to-3D synthesis, it often introduces artifacts such as over-smoothed geometry, cartoonish and vague textures, and even the multi-face Janus problem. 
Additionally, these methods struggle to accurately interpret fine-grained details from text prompts, and cannot process long descriptions due to the length limit of the text encoder. A more fundamental challenge in text-to-human generation lies in the entanglement of human body and clothing. Such entanglement not only makes it difficult to customize or exchange garments without full avatar regeneration but also hinders the implementation of physically based simulations for realistic apparel behavior. Critically, prevalent neural 3D representations like Neural Radiance Fields (NeRF)~\cite{mildenhall2021nerf} or 3D Gaussian Splatting~(3DGS)~\cite{3DGS}, despite their impressive novel view rendering capabilities~\cite{DBLP:conf/nips/KolotourosAZBFS23,HumanGaussian}, lack inherent compatibility with traditional computer graphics (CG) pipelines, which predominantly utilize polygonal meshes for human body and strands for hairs. This incompatibility hinders their integration into established workflows for character animation, physical cloth simulation, and production rendering.

Concurrently, text-to-3D-garment generation faces its own distinct set of hurdles. Approaches based on sewing patterns~\cite{he2024dresscode}, while offering a structured garment representation, often overfit to training datasets and templates, resulting in limited output diversity and poor adherence to input textual prompts.
Moreover, many existing works also overlook body shape adaption~\cite{li2025garmentdreamer,sarafianos2024garment3dgen}, often necessitating post-generation adjustments and reducing immediate ``out-of-the-box'' usability.
Furthermore, achieving realistic and detailed garment textures remains challenging, due to unresolved technical barriers in handling self-occlusion and possibly preserving texture symmetries during the generation process.

To address these multifaceted challenges and meet the growing demand for production-ready 3D character assets, we introduce {\it Tailor} --- an integrated text-to-3D framework that generates CG-ready avatars with customized garments (see Fig.~\ref{fig:teaser}). In our study, we use the term {\it CG-ready} for models that adhere to a technical aesthetic common in computer graphics (CG) and adopt CG-compatible representations such as polygonal meshes and hair strands --- as opposed to neural representations like NeRF and 3DGS. This design ensures seamless integration into conventional CG pipelines and support downstream tasks such as physical simulation. Tailor uniquely proposes a generalizable methodological framework that combines LLM-guided procedural human generation with a novel, training-free garment synthesis pipeline, forming a cohesive, customizable and extensible system.
In particular, we make the following contributions in this paper:
\begin{itemize}
\item {\it An Integrated Text-to-Avatar Framework.} We introduce Tailor, an integrated system that generates both high-fidelity customizable 3D avatars and physically compatible garment (ready for downstream simulation tasks) from textual descriptions. Unlike existing methods that treat avatar and garment creation separately, our approach unifies both processes into a coherent framework, significantly improving efficiency and interoperability. As a system-level methodology, Tailor also brings unique insight to end-to-end automated CG-ready assets generation.
\item {\it Novel Technical Components.} Tailor incorporates three key innovations. (1) Semantic-Aware Human and Garment Parsing: we leverage an LLM agent to interpret free-form textual descriptions into detailed parametric human models and semantically matched garment templates, allowing intuitive and detailed customization. (2) Geometry-Aware Garment Generation: we propose a novel geometry-aware deformation process that generates garments under text control while preserving mesh topology and incorporating geometric constraints for body alignment, realism and simulation readiness. (3) Multi-View Consistent Texture Synthesis: we develop a diffusion process tailored for garment texture generation, ensuring multi-view consistency, high visual fidelity, and optional symmetry depending on garment design.
\item {\it Comprehensive Empirical Validation.} We perform comprehensive qualitative and quantitative experiments, demonstrating that Tailor outperforms existing state-of-the-art methods. Tailor effectively addresses critical limitations in existing methods such as 3D data scarcity and extensive manual effort, providing a amateur-friendly solution that efficiently produces high-quality, ready-to-use avatars from text.
\end{itemize}

\section{Related Work}

\subsection{Text-to-3D Human Generation}

Generating 3D humans from text has rapidly evolved, primarily leveraging pretrained vision-language models (e.g., CLIP~\cite{clip}) and 2D diffusion models (e.g., text-to-image diffusion models~\cite{SDS}) while incorporating human body priors~\cite{SMPL:2015, SMPL-X:2019}. AvatarCLIP~\cite{hong2022avatarclip} optimizes an SMPL-based NeRF under CLIP guidance, while DreamHuman~\cite{DBLP:conf/nips/KolotourosAZBFS23} proposes to use SDS~\cite{SDS} to learn a deformable, pose-conditioned NeRF for animatable human. HumanGaussian~\cite{HumanGaussian} replaces NeRF with 3DGS anchored to an SMPL-X model, enabling faster training and rendering. TADA~\cite{DBLP:conf/3dim/LiaoYXTHTB24} proposes an optimizable human body model built upon the displacement-enhanced SMPL-X and a texture map, explicitly targeting CG pipeline compatibility. DreamWaltz-G \cite{DreamWaltz-G} utilizes a hybrid 3D Gaussian avatar representation and improves animatability by skeleton-guided score distillation. However, these works generally focus on learning holistic representations which entangle the body and garments, thereby restricting clothing editing and simulation.

To overcome this limitation, several recent works have focused on disentangling clothing from the body. AvatarFusion~\cite{DBLP:conf/mm/HuangYLY023} proposes a post-processing optimization for clothing separation. More integrated approaches represent garments as distinct layers: SO-SMPL~\cite{SO-SMPL} sequentially deforms SMPL-based body and clothing meshes; TELA~\cite{tela} proposes layer-wise NeRF-based disentangled generation; Barbie~\cite{barbie} uses a multi-stage pipeline with expert models on SMPL-based DMTets~\cite{shen2021deep} for disentangled generation; and SimAvatar~\cite{SimAvatar} adopt hybrid 3D representations---SMPL mesh, Unsigned Distance Field (UDF)~\cite{MeshUDF}, and hair strands---to model the body, garment, and hair, respectively. While these methods successfully separate the body and apparel, generating highly realistic avatars and garments remains challenging. In particular, a notable quality gap persists in the generated garments when compared with specialized state-of-the-art text-to-garment methods.
 
Our work, Tailor, addresses these challenges through a hybrid strategy that decouples the generation of the human body from the garment. For the body, we adopt an LLM-guided procedural approach that produces high-quality, fully animatable avatars (including facial expressions) with strong prompt adherence and CG compatibility. This sidesteps the speed and fidelity limitations of distillation-based avatar pipelines, which often yield distorted faces, irregular body shapes, and oversaturated textures.
The decoupled stages and representations also enables us to dedicate a separate and more powerful generative pipeline for garment generation, achieving quality competitive with specialized text-to-garment methods.

\subsection{Text-to-3D Garment Generation}

\begin{figure*}[ht]
\centering
\includegraphics[width=1\linewidth]{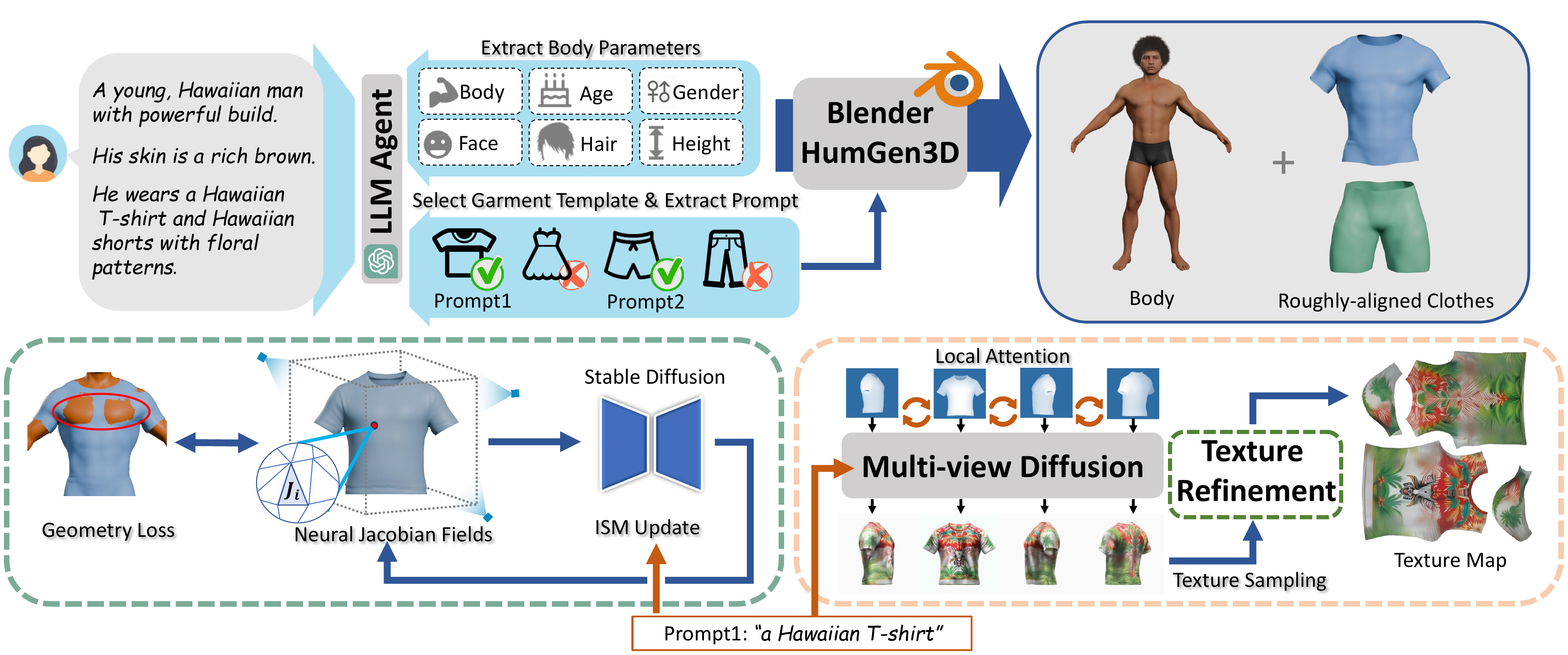}
\caption{\textbf{Overview of Tailor.} Tailor includes a three-stage pipeline. Given a description of a clothed human, 
(a) an LLM agent decomposes the prompt into separate body and garment sub-prompts, then outputs translated body parameters and garment templates. The body parameters and garment templates are subsequently fed into the HumGen3D generator to derive a highly-detailed human model and a set of roughly-aligned garments.
(b) The framework applies topology-preserving deformation to the Neural Jacobian Field of garment template under the guidance of both a text-to-image model and geometry constraints to generate body-aligned clothes.
(c) We condition a multi-view image diffusion model with rendered depth images to generate view-consistent images by introducing a local attention mechanism. Then, textures are sampled from the images and refined to get the final texture map.} \label{fig:pipeline}
\end{figure*}

\textbf{Garment Geometry Generation.} Current approaches generally follow two main paradigms: pattern-based construction or direct 3D synthesis. Pattern-based methods offer inherent structure and simulatability by generating 2D sewing patterns. DressCode~\cite{he2024dresscode} tokenizes garment patterns and employs a GPT-style model to autoregressively generate these token sequences from a text prompt. ChatGarment~\cite{ChatGarment} uses a Vision-Language Model (VLM) to interpret text or image inputs into a structured JSON file that programmatically defines patterns. AIpparel \cite{nakayama2024aipparellargemultimodalgenerative} trains a multimodal autoregressive model to generate sewing tokens. While structured, these methods are constrained by strong inductive biases and limited training data diversity. In comparison, direct 3D synthesis methods offer more geometric freedom. GarmentDreamer~\cite{li2025garmentdreamer} optimizes 3DGS~\cite{3DGS} via SDS~\cite{SDS}, but it struggles to recover smooth surfaces critical for garment modeling and simulation. Garment3DGen~\cite{sarafianos2024garment3dgen} first reconstructs a pseudo-3D target from either a real or text-generated image, and then deforms a garment template to fit this target. However, the reliance on the intermediate pseudo-3D reconstruction often introduces inaccuracies, such as unnaturally flattened surfaces and erroneous watertight meshes.
Surf-D~\cite{Surf-D} explores 3D garment generation by applying a diffusion model~\cite{DDPM} to UDF~\cite{MeshUDF}.
However, many of these approaches lack of body awareness; they often fail to generate garments that are aligned with different body shapes and poses, hindering immediate ``out-of-the-box'' usability. Tailor addresses these challenges by deforming garment templates under a powerful text-to-image model and novel geometric constraints, ensuring high-quality, diverse, and body-aligned geometry.

\textbf{Garment Texture Generation.} Generating consistent and realistic garment textures remains challenging due to complex clothing properties like non-watertight meshes, self-occlusions, and possible symmetry constraints. Garment3DGen~\cite{sarafianos2024garment3dgen}  uses a ``project-and-inpaint'' approach, which 
often introduces multi-view inconsistencies due to limited global coherence. DressCode~\cite{he2024dresscode} sidesteps this by finetuneing a text-to-image diffusion model for 
Physically Based Rendering (PBR) materials, but it is limited to simple repetitive patterns. GarmentDreamer~\cite{li2025garmentdreamer} employs SDS for more diverse outputs, yet often inherits distillation artifacts such as blurring and oversaturation.

Meanwhile, general mesh texturing has shifted from to ``project-and-inpaint'' paradigm~\cite{richardson2023texture, chen2023text2tex, zeng2024paint3d} to consistent multi-view diffusion strategies~\cite{SyncMVD, lu2024genesistex2, Hunyuan3D-2}, demonstrating improved cross-view consistency. Paint3D~\cite{zeng2024paint3d} adopts a two-stage pipeline: a coarse stage combining depth-based generation with inpainting, followed by UV-space diffusion to refine textures. However, it still suffers from multi-view inconsistencies, and its generalization is limited by the scale and diversity of available 3D texture datasets. SyncMVD~\cite{SyncMVD} introduces a synchronized multi-view diffusion framework that employs an aggregate-and-render mechanism alongside cross-view attention to enforce cross-view consistency. GenesisTex2~\cite{lu2024genesistex2} proposes a local attention reweighting mechanism based on geometric distance to further enhance the fidelity of local details. Hunyuan3D-2~\cite{Hunyuan3D-2} generates a frontal reference image first and then synthesizes multi-view mesh images conditioned on geometry. Nonetheless, these approaches do not directly address the unique challenges posed by garments. To overcome these limitations, we introduce a series of modifications to existing multi-view diffusion pipelines, specifically tailored for garment texturing, to achieve consistent and high-fidelity results.

\section{Method}

Our framework includes a three-stage pipeline (See Fig.~\ref{fig:pipeline}). Given a textual description of a clothed human, we first employ a large language model to generate a 3D human body and sematically match garment templates (Sec.~\ref{subsec:hum_gen_llm}). Then, we adopt topology-preserving deformation to the selected garment templates, under the supervision of both a text-to-image  model and body-aligned geometric constraints.
(Sec.~\ref{subsec:garment_geo_gen}). Finally, we utilize a specially designed multi-view diffusion module to generate high-fidelity textures for the generated garments (Sec.~\ref{subsec:texture}).

\subsection{Preliminaries}

We employ HumGen3D\footnote{\url{https://www.humgen3D.com/}}---a professional 3D human generator add-on for Blender%
, as the foundation of our pipeline. We strategically favor this parametric human representation over other neural representation or 3DGS based end-to-end generation method to ensure production-grade asset quality. While these generation approaches often suffer from geometric and textural artifacts, our use of a human prior guarantees valid topology, precise rigging, and compatibility with standard graphics workflows.

Specifically, HumGen3D represents the human body as a textured blendshape model, controlled by a comprehensive set of hierarchical parameters (e.g., age, body proportions, skin tone, and facial features). In addition, the add-on tool provides a variety of strand-based hairstyles and basic clothing presets. Despite these robust capabilities, creating detailed human models---especially those with intricate clothing variations---requires considerable artistic expertise, time-consuming adjustments, and extensive manual intervention. To overcome these challenges, we fully automate the human and garment modeling process, using LLM agents integrated within a 3D generation pipeline. Our method enables even novice users to effortlessly produce high-fidelity, diverse, and richly customizable clothed human avatars.

\subsection{Human Generation and Template Matching} \label{subsec:hum_gen_llm}

We first construct a garment template library
by combining HumGen3D’s built-in garment presets with additional garment meshes sourced from the Internet. The library is organized into three categories: \textit{upper-body} (e.g., short/long-sleeve tops and sleeveless tops), \textit{lower-body} (e.g., trousers, shorts, and skirts), and \textit{full-body} (e.g., dresses and jumpsuits). Each garment template is rig-bound to base human skeletons through automated skinning weights, enabling automatic adaptation to diverse body shapes.

To generate a 3D human model and accurately match garment templates from textual input, we propose an LLM-driven workflow with three key steps, as shown in the top row of Fig.~\ref{fig:pipeline}: 
\begin{itemize}
    \item \textbf{Prompt Decomposition:} The input description encompasses both physical attributes and garment specifications. The LLM agent decomposes the overall prompt into distinct body and garment sub-prompts, enabling granular subsequent processing. 
    \item \textbf{Body Parameterization:} The LLM agent translates the body-related sub-prompt into body parameters, and programmatically invokes HumGen3D’s API for automatic human body generation and hairstyle selection from the asset library. \item \textbf{Garment Template Matching:} The LLM agent retrieves optimal garment templates from the garment library for each user-specified garment based on the garment sub-prompt, and decides whether symmetry is presented in geometry and texture.
\end{itemize}

To enable the LLM to perform these tasks effectively, we leverage in-context learning techniques by providing the LLM with parameter descriptions, valid ranges, template descriptions, and annotated examples. Further details of the LLM agent are provided in the  Appendix. Finally, a highly detailed 3D human model $H$ and roughly-aligned garment meshes $\{M_{tpl}\}_i$ are generated from HumGen3D generator.

\subsection{Body-Aligned Garment Geometry Generation} \label{subsec:garment_geo_gen}

\begin{figure}[htbp]
\centering
\includegraphics[width=\linewidth]{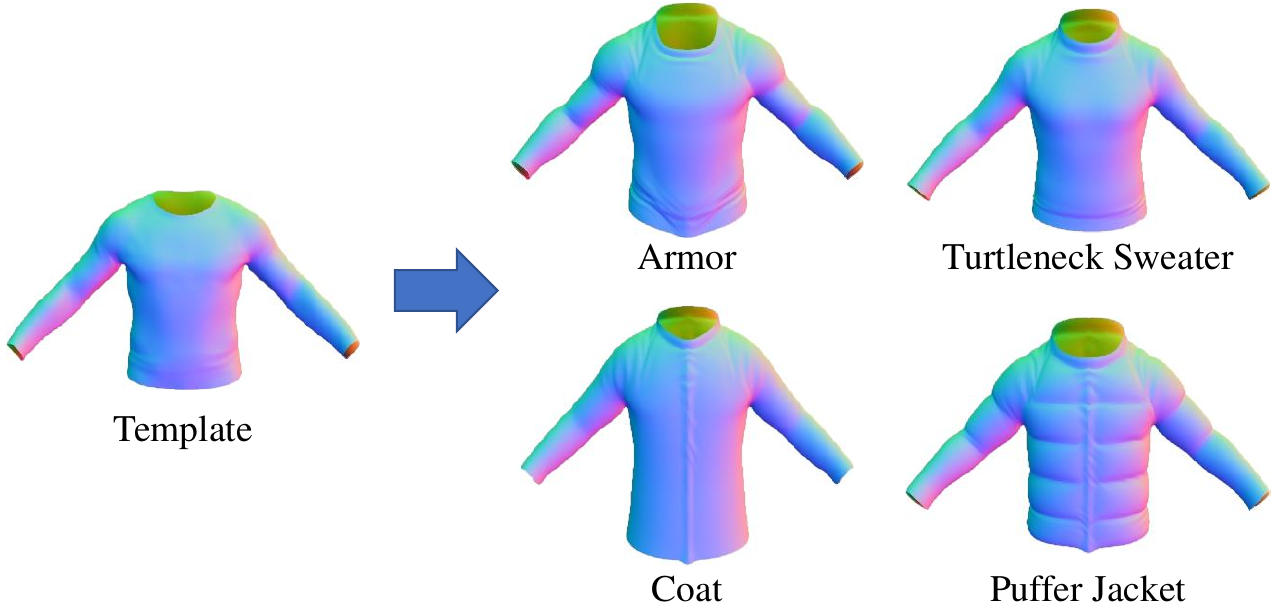}
\caption{Starting from the same template, our topology-preserving deformation process generates diverse and plausible garment geometries conditioned on different target prompts.
}
\label{fig:deformation}
\end{figure}

In this stage, we generate body-aligned garments based on the garment sub-prompts provided by the LLM agent described in the last section. Given a garment template mesh $M_{tpl}$ and its corresponding sub-prompt $y$, we deform $M_{tpl}$ under the guidance of both a text-to-image model~\cite{SD3} and a set of geometry losses through Neural Jacobian Fields~(NJF)~\cite{aigerman22neuraljacobianfields}. As shown in Fig.~\ref{fig:deformation}, our approach enables the creation of diverse and plausible garment geometries from only a small set of templates.

\textbf{Text-Driven Supervision.} 
We optimize the NJF $J$ of the template mesh $M_{tpl}$, guided by a text-to-image Rectified Flow model~\cite{SD3, RectifiedFlow}. Specifically, we derive the deformation map $\Phi^*$ from $J$, and use a differentiable renderer~\cite{nvdiffrast} to render the normal map $n = g(\Phi^*(M_{tpl}), c)$ from a randomly sampled camera view $c$ in the spherical coordinate system. This normal map is then 
encoded into the latent representation $x$ of the image model. To optimize NJF, we adapt Interval Score Matching (ISM)~\cite{LucidDreamer} to Rectified Flow and derive the corresponding gradient (expectation notation is omitted for brevity):
\begin{equation}
    \nabla_J \mathcal L_{ISM} = w(t) \left[v_{\phi}(x_t, t, y) - v_{\phi}(x_s, s, \emptyset) \right] \frac{\partial g}{\partial J},
\end{equation}
where $x_t$ and $x_s$ are deterministically computed intermediate latents at timesteps $t$ and $s$ along the rectified flow trajectory. $v_{\phi}(x_t, t, y)$ denotes the rectified flow’s vector field conditioned on the text prompt $y$ at timestep $t$. By leveraging deterministic trajectories and interval-based gradients, the adapted ISM avoids the inconsistent gradient signals characteristic of traditional SDS. This ensures stable optimization updates, preserves geometric details, and accelerates convergence toward a high-fidelity and realistic final 3D clothing geometry.

\textbf{Body-Alignment Geometry Supervision.} The garment templates serve as the
initialization for generation.
However, they are only roughly aligned to the body by automatic skinning, which may result in garment-body penetrations (as illustrated in the lower-left corner of Fig.~\ref{fig:pipeline}). Moreover, due to the lack of body geometry awareness, the text-guided deformation process can introduce additional issues such as further penetrations, disproportionate scaling, and self-intersections.
To mitigate this, we propose a Body Collision Loss $\mathcal L_{coll}$.
In each iteration, we uniformly sample $N_p$ points $\{p_i\}$ from the garment surface and calculate their signed distance $d_{body,i}$ to the human body mesh $H$. The loss is defined as:
\begin{equation}
    \mathcal L_{coll} = \mathbb E_i \left[ \max(\epsilon - d_{body,i}, 0) \right],
\end{equation}
where $\epsilon$ is a small positive number. 

\begin{figure}[htbp]
\centering
\includegraphics[width=0.3\linewidth]{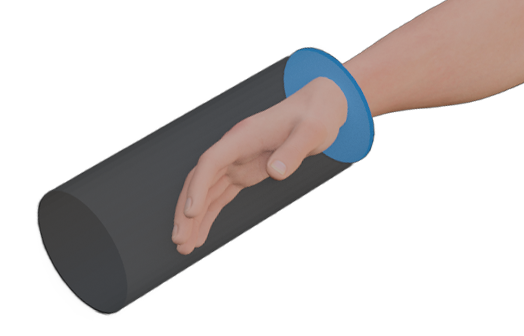}
\caption{The predefined semi-infinite cylinders for $\mathcal L_{blk}$.
}
\label{fig:cylinder}
\end{figure}

In addition, the deformation may occasionally lead to positional shifts or disproportionate scaling during optimization, causing misalignment such as sleeves extending beyond the wrists or the waistline being positioned too high. We introduce a blocking loss $\mathcal L_{blk}$ to address this issue. As illustrated in Fig.~\ref{fig:cylinder}, leveraging the consistent topology in HumGen3D-generated human meshes, we predefine a set of semi-infinite cylinders with one closed end near the joints (e.g., wrists, ankles, neck, and waist), and penalize sampled garment surface points that fall within these cylinders by computing their distance $d_{end,i}$ to the closed end. The loss is formulated as:
\begin{equation}
    \mathcal L_{blk} = \mathbb E_i \left[ d_{end,i} \cdot \mathbbm{1}(p_i\text{ inside cylinder}) \right].
\end{equation}
This simple loss effectively restricts implausible garment deformation and prevent positional shifts and disproportionate scaling during optimization.

\begin{figure}[htbp]
\centering
\includegraphics[width=\linewidth]{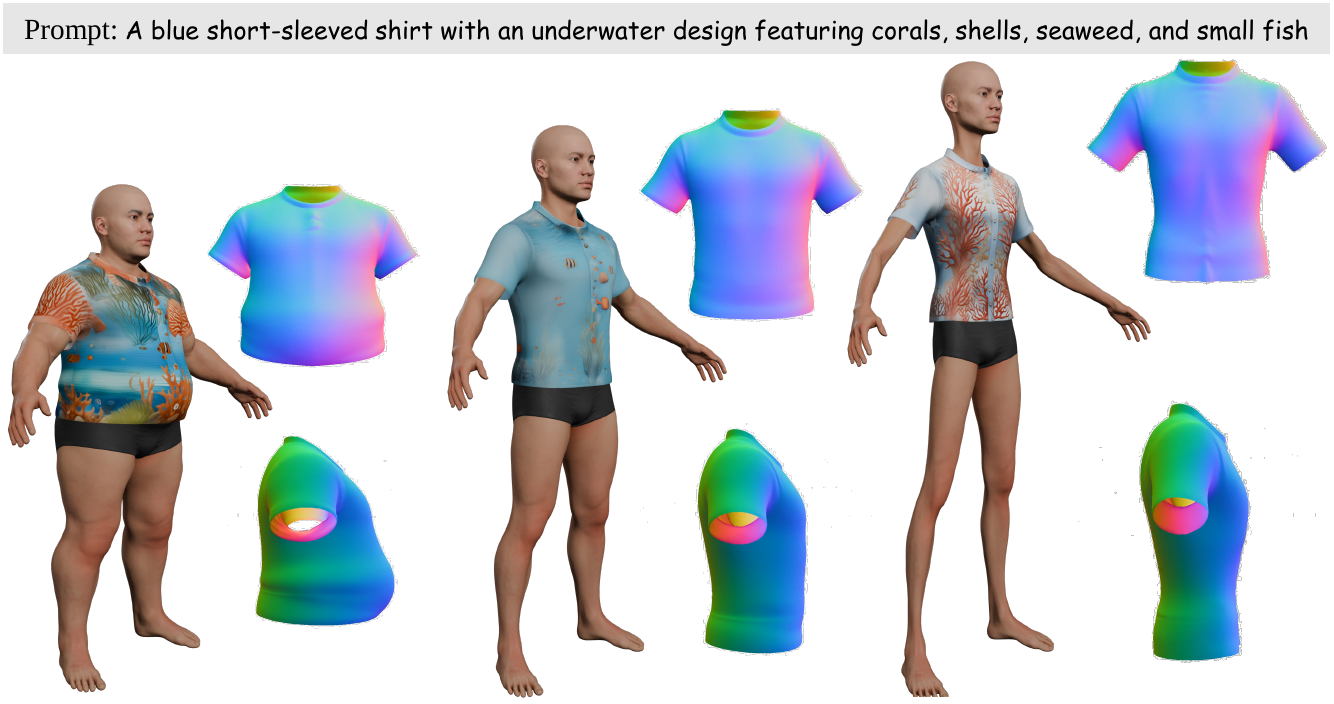}
\caption{Given the same template and text prompt, our method can generate body-fitting clothing for people of different body shapes.
}
\label{fig:body_shape}
\end{figure}

Our template initialization, along the with carefully designed body collision loss $\mathcal L_{coll}$ and blocking loss $\mathcal L_{blk}$, enables our method to create well-fitting garments for people of different body shapes, as shown in Fig.~\ref{fig:body_shape}.

\textbf{Symmetry Supervision.} Since garments often exhibit bilateral symmetry, we introduce an optional symmetry loss $\mathcal{L}_{sym}$.
We reflect the sampled points $\{p_i\}$ across the X-axis to obtain a mirrored set $\{ p^\prime_i \}$. The symmetry loss is then defined as the Chamfer distance between these two sets:
\begin{equation}
    \mathcal{L}_{sym} = \frac{1}{N_p} \sum_{i} \min_{j} \| p_i - p^\prime_j \|_2^2 + \frac{1}{N_p} \sum_{j} \min_{i} \| p^\prime_j - p_i \|_2^2.
\end{equation}

\textbf{Regularizations.} To encourage the generation of evenly-distributed faces and smooth surface, which are crucial for simulation, we adopt two regularization terms, including a Laplacian loss $\mathcal{L}_{lap}$ and a normal consistency loss $\mathcal{L}_{nc}$~\cite{worchel22multi}.

\textbf{Overall Loss.} The overall loss for body-aligned garment geometry generation can be summarized into:
\begin{equation}
\begin{split}
    \mathcal L_{geo} &= \mathcal L_{ISM} + \lambda_{coll} \mathcal L_{coll} + \lambda_{blk} \mathcal L_{blk} \\
    &\quad + \lambda_{sym} \mathcal{L}_{sym} + \lambda_{lap} \mathcal L_{lap} + \lambda_{nc} \mathcal L_{nc}.
\end{split}
\end{equation}

\subsection{Texture Generation} \label{subsec:texture}

We propose a multi-view diffusion-based approach for synthesizing high-fidelity 2K-resolution textures on garment meshes that align with input text prompts.

\textbf{Synchronized Multi-View Diffusion.} Given the deformed garment mesh from the preceding stage, we render depth maps from $N_v$ surrounding viewpoints. These depth maps condition a text-to-image diffusion model~\cite{SDXL} via ControlNet~\cite{ControlNet}. The generated multi-view images are then back-projected onto the UV texture space to produce an aggregated texture map. 
However, this approach suffers from multi-view inconsistencies due to independent per-view diffusion processes, resulting in fragmented textures and visible seams. To address this, we integrate 
a latent merging pipeline and cross-view attention sharing mechanisms inspired by~\cite{SyncMVD, GenesisTex2}.

The latent merging pipeline operates within the latent space of an image diffusion model to maintain 3D consistency across multiple views. The process firstly initializes a latent texture $W_T$ by sampling from a standard normal distribution, and then projects it into image space to create 3D-consistent initial noisy views $Z_T = \{ z_T^{(v_i)} \}_{i=1}^{N_v}$ aligned with the input mesh. During denoising at timestep $t$, while per-view noiseless estimates $Z_{0|t}$ lack multi-view coherence because of independent predictions, the pipeline addresses this by: 1) Back-projecting each $Z_{0|t}$ view into the shared texture space to form partial texture maps, 2) Aggregating the partial maps into a unified latent texture $W_{0|t}$, and 3) Reprojecting the consolidated texture into image space to update $Z_{0|t}$. This cyclic merging enforces geometric consistency across views throughout denoising iterations.

The core of the latent merging pipeline is the aggregation operation:
\begin{equation}
    W_{0|t} = \frac{\sum_{i=1}^{N_v} \alpha^{(v_i)} \cdot \text{backproj}(z_{0|t}^{(v_i)})}{\sum_{i=1}^{N_v} \alpha^{(v_i)}},
\end{equation}
where $\alpha^{(v_i)}$ is the weight for the view $v_i$, which is determined by the angle between the normal vector of the 3D point and the reversed view direction~\cite{GenesisTex2}.
However, this operation often causes loss of high-frequency details, resulting in over-smoothed or blurry outcomes. We further observe that garment texture details mainly comes from the front and back views. To prioritize the texture in these key regions, we update the non-front and non-back 
$\alpha^{(v_i)}$ with 
\begin{equation}
    \alpha^{(v_i)}_{new} = \left[1-\left(\alpha^{(v_{front})}+\alpha^{(v_{back})}\right)/\sum_{j=1}^{N_v} \alpha^{(v_j)}\right] \alpha^{(v_i)},
\end{equation}
which adaptively downweights the contributions from side views, thereby allocating more capacity to preserve fine-grained textures from the front and back views without discarding multi-view consistency.

\begin{figure}[ht]
\centering
\includegraphics[width=0.85\linewidth]{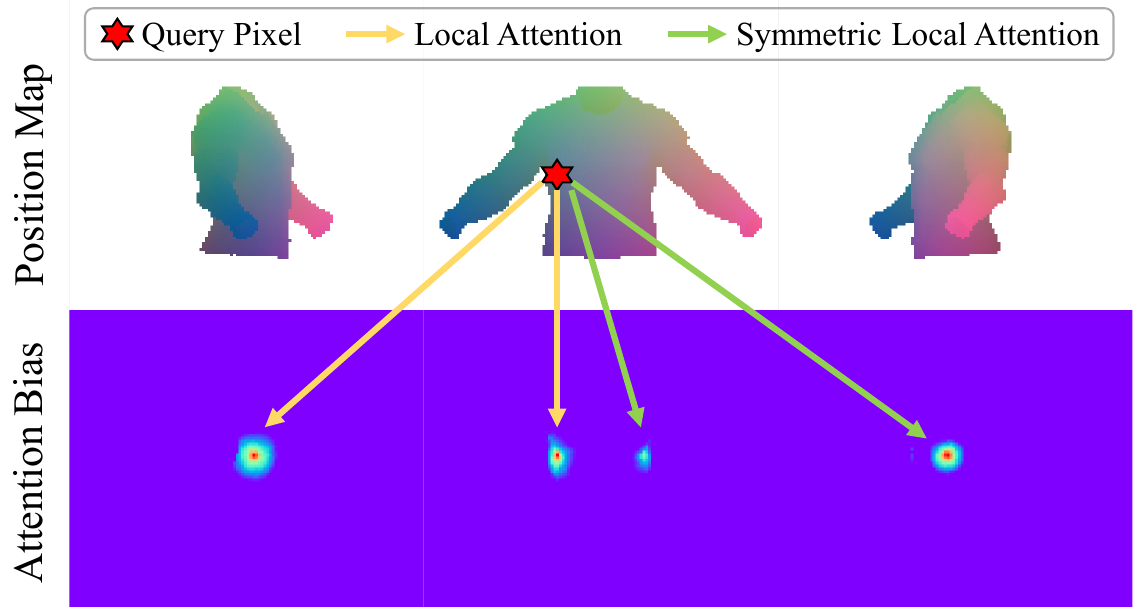}
\caption{Illustration of the local attention and symmetric local attention.} \label{fig:sym_attn}
\end{figure}

\textbf{Symmetric Local Attention.} While cross-view attention adaptations~\cite{SyncMVD} of Stable Diffusion's self-attention improve multi-view consistency, they often compromise color diversity and local detail fidelity. 
To overcome this limitation, we introduce the geometry-aware local attention bias matrix from~\cite{GenesisTex2}. This matrix recalibrates attention weights by prioritizing pixels that share underlying 3D surface points across views, computed via Euclidean distances between corresponding 3D coordinates. Since garments frequently exhibit symmetric texture patterns, we propose to augment this mechanism by simultaneously evaluating distances to X-axis mirrored 3D points. 
The resulting bias matrix emphasizes both direct correspondences and their symmetrical counterparts, as illustrated in Fig.~\ref{fig:sym_attn}. This dual-distance metric serves as an optional prior that enforces bilateral symmetry in texture generation while maintaining local detail retention capabilities. The degree of symmetry enforcement is adjustable through relative weighting of original
and mirrored distance contributions in the bias calculation.

\begin{figure}[ht]
\centering
\includegraphics[width=0.92\linewidth]{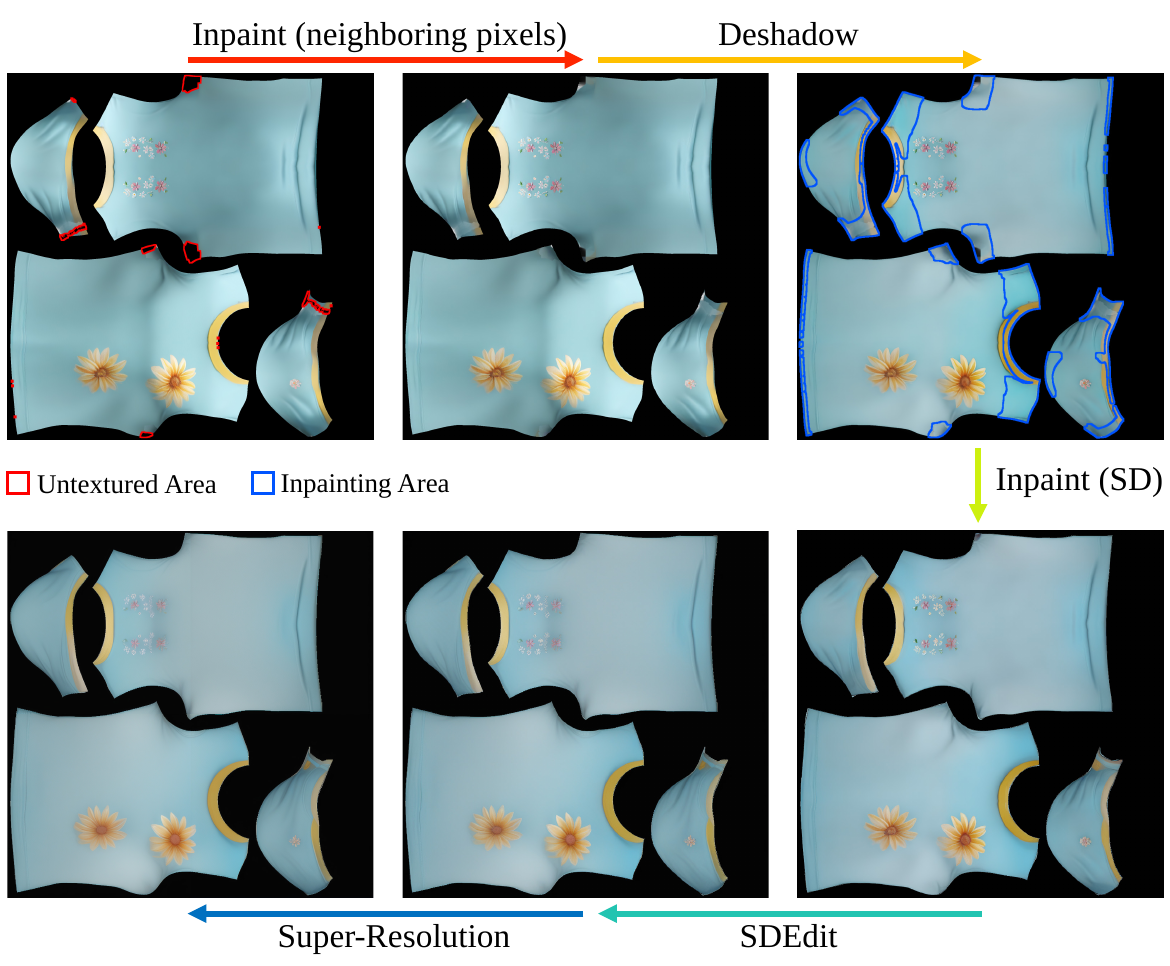}
\caption{The texture refinement pipeline.} \label{fig:tex_refine}
\end{figure}

\textbf{Texture Refinement.} While synchronized multi-view diffusion yields globally coherent texture maps, challenges such as self-occlusions, baked shadows, and inconsistent lighting can introduce localized artifacts, including texture holes, blurry regions, and illumination biases. 
To address these, we apply a UV-space refinement pipeline as illustrated in Fig.~\ref{fig:tex_refine}. This approach exploits the pre-defined UV parameterization of our garment templates, which are constructed as sewing-pattern layouts composed of large, spatially continuous, and semantically coherent regions. This structural resemblance to flattened clothing images ensures that models trained on natural images can be effectively utilized for refinement.
We begin by performing preliminary inpainting to fill untextured voids using valid neighboring pixels.
For non-metallic materials, we then use a pretrained shadow removal network~\cite{ShadowR} to eliminate baked shadows and lighting artifacts from the texture map. Subsequently, we employ Stable Diffusion XL~\cite{SDXL} to inpaint regions that have limited visibility in the original multi-view images. In cases of pronounced self-occlusion, such as garments with long sleeves, we further refine the texture maps using SDEdit~\cite{SDEdit}, guided by texture masks. Finally, we upsample the texture resolution from 1K to 2K using UniControlNet~\cite{UniControlNet}. 

\section{Experiment}

\begin{figure*}[tbp]
\centering
\includegraphics[width=1\linewidth]{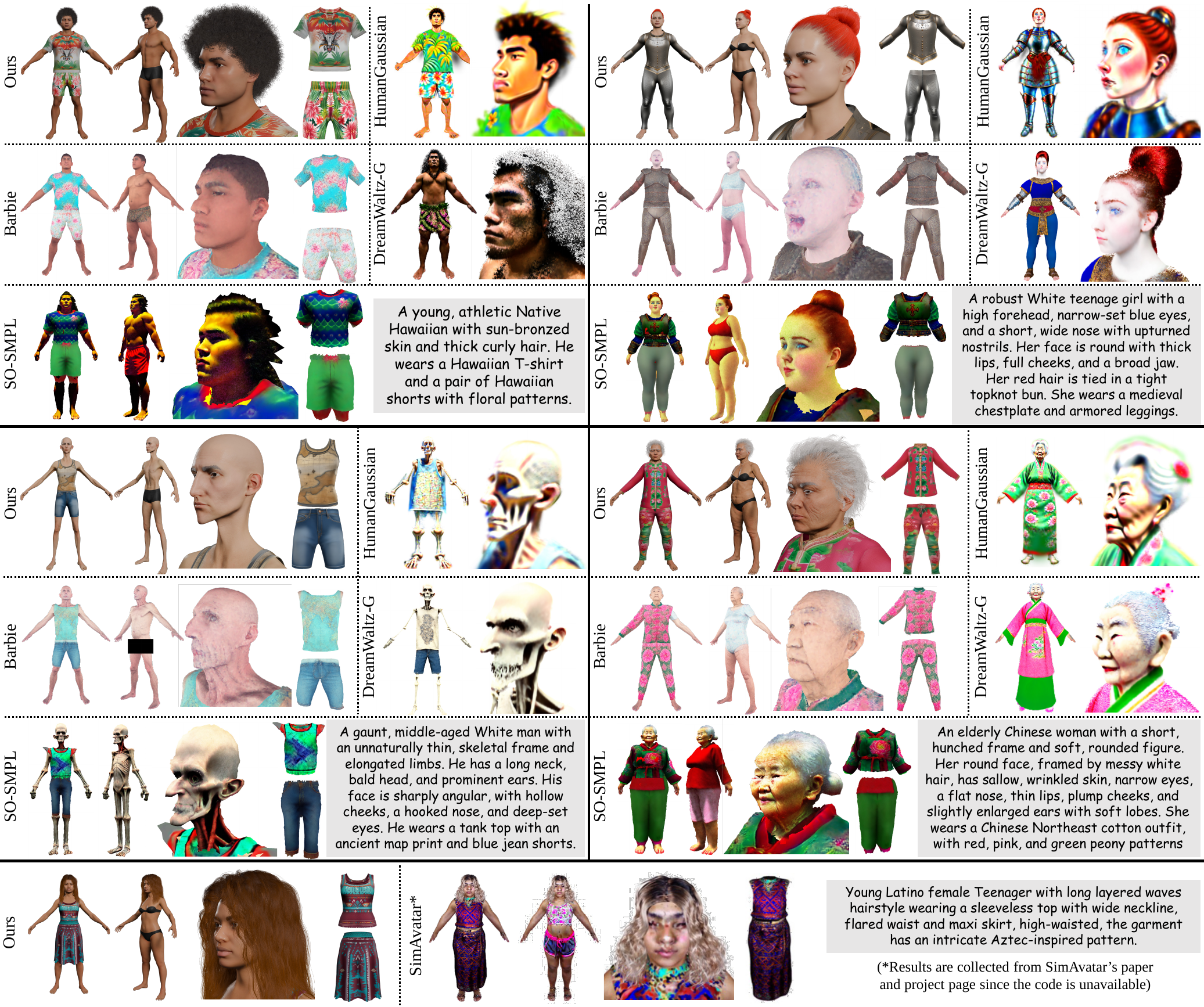}
\caption{Qualitative comparison with SoTA text-to-3D human generation methods. Our method consistently generates high-fidelity 3D clothed humans that demonstrate superior details and alignment with text prompts. In contrast, competing methods often struggle with semantic accuracy, produce lower-quality textures, or lack photorealism.
}
\label{fig:t2h-1}
\end{figure*}

\begin{figure*}[htbp]
\centering
\includegraphics[width=0.98\linewidth]{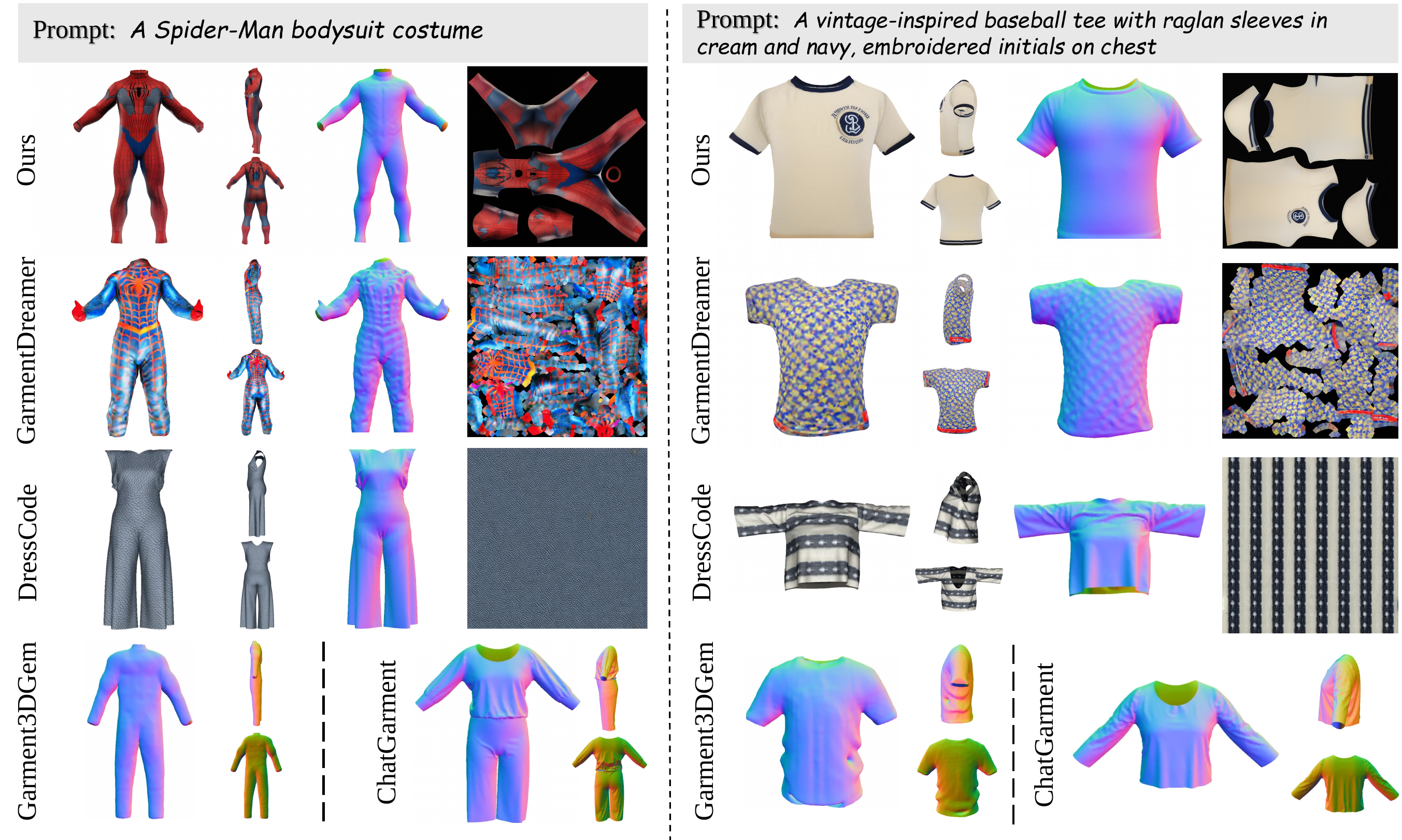}
\caption{Qualitative comparison with state-of-the-art text-to-3D garment generation methods. We show the rendered RGB image, normal map, and texture map for each method. For DressCode which generates PBR texture, we show the diffuse component.
}
\label{fig:t2g-1}
\end{figure*}

\begin{figure*}[htbp]
\centering
\includegraphics[width=0.99\linewidth]{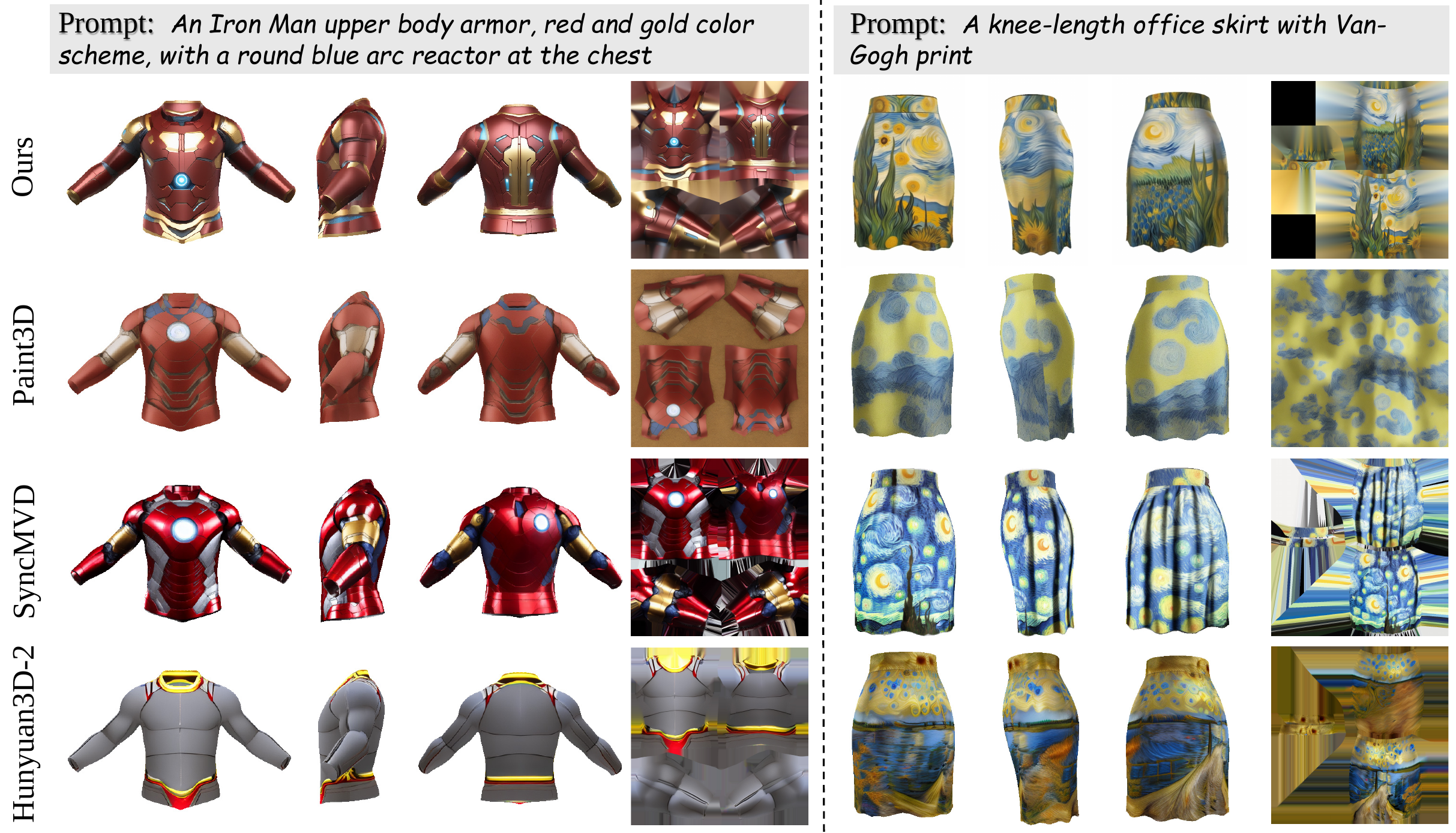}
\caption{Qualitative comparison with state-of-the-art mesh texturing methods.
Our texturing pipeline demonstrates a superior ability to generate high-fidelity textures that are semantically and stylistically faithful to text prompts. Note our model's success in rendering an iconic character with precise details (the ``Iron Man'' armor and ``arc reactor'') and a recognizable artistic style (``Van Gogh print'').
}
\label{fig:comparison_texture}
\end{figure*}

\subsection{Implementation Details}

Our LLM agent is based on OpenAI GPT-4o~\cite{GPT4o}. 
For garment geometry generation, we set  $\lambda_{coll}=5\mathrm{e}5$, $\lambda_{blk}=1\mathrm{e}5$, $\lambda_{lap}=\lambda_{nc}=2\mathrm{e}4$, and $\lambda_{sym}=5\mathrm{e}5$ (optional for garments with symmetric geometry). The $\epsilon$ for $\mathcal L_{coll}$ is set to $0.005$. Garment coordinates are normalized to $[-1, 1]$, and $N_p = 50000$ points are sampled per iteration.
We employ Stable Diffusion 3.5~\cite{SD3} as the text-to-image model and optimize the NJF with Adam~\cite{Adam}, using a learning rate of $0.002$, for 600 iterations and a batch size of 4.
We sample the diffusion timestep $t$ from $[500, 980]$ for the first 300 iterations and $[50, 980]$ for the remaining iterations. $w(t)=1$ for all $t$. 
For texture generation, we employ Stable Diffusion XL~\cite{SDXL} to generate $N_v=6$ views along the equator surrounding of the garment. 
Our method achieves text-to-3D clothed human generation in 12 minutes 40s on a single A100 GPU, with 10s for human generation and additional time allocated for garment geometry, texture, and refinement stages.

\subsection{Evaluation Metrics}

We render the generated 3D human or garment and evaluate the following metrics: \textbf{a) CLIP and FashionCLIP (FCLIP) Score.} We measure the semantic alignment between the rendered images and the text prompts using CLIP Score~\cite{clip}. For garments, we additionally use FashionCLIP~\cite{fashionclip}, a CLIP model fine-tuned on fashion data for domain-specific evaluation. \textbf{b) Aesthetic Score~(Aesth.).} The visual appeal of the generated images is quantified using a SigLIP-based~\cite{SigLIP} aesthetic predictor\footnote{\url{https://github.com/discus0434/aesthetic-predictor-v2-5}}, which provides a score for aesthetic quality. \textbf{c) Image Quality~(IQ).} We use the no-reference image quality assessment model MANIQA~\cite{MANIQA} to evaluate the perceptual quality of the generated images.

\subsection{Comparisons of Text-to-3D Human Generation}

\begin{table}[htb]
    \renewcommand{\arraystretch}{1.1}
    \caption{Quantitative comparison for text-to-3D human generation.}
    \label{tab:t2h_metrics}
    \centering
    \begin{tabular}{l|ccc|cc}
    \Xhline{2\arrayrulewidth}
    \multirow{2}{*}{Method} & \multicolumn{3}{c|}{Quantitative Metrics} & \multicolumn{2}{c}{Avg. User Ranks} \\
     & CLIP & Aesth. & IQ & TA & VQ \\
    \Xhline{2\arrayrulewidth}
    SO-SMPL & 24.35 & 2.89 & 0.509 & 2.77 & 2.76 \\
    HumanGaussian & 25.28 & 3.23 & 0.191 & 3.17 & 3.69 \\
    DreamWaltz-G & 23.91 & 3.52 & 0.357 & 3.36 & 3.53 \\
    Barbie & 24.83 & 3.45 & 0.498 & 3.76 & 3.68 \\
    Ours & \textbf{26.52} & \textbf{4.11} & \textbf{0.595} & \textbf{1.93} & \textbf{1.34} \\
    \Xhline{2\arrayrulewidth}
    \end{tabular}
\end{table}

We compare Tailor with the state-of-the-art text-to-3D human generation methods, categorized into holistic methods (HumanGaussian~\cite{HumanGaussian}, DreamWaltz-G~\cite{DreamWaltz-G}) and disentangled methods (SO-SMPL~\cite{SO-SMPL}, SimAvatar~\cite{SimAvatar}, and Barbie \cite{barbie}). We show qualitative comparisons in Fig.~\ref{fig:t2h-1} and 
Fig.~\ref{fig:t2h-2}. 
For quantitative evaluation, we generate 10 distinct clothed human models from text prompts that span a range of genders, ages, ethnicities, physical appearances, and clothing styles, and render them from various shot types and angles.
Metrics are computed across all rendered views, and the results are summarized in Table~\ref{tab:t2h_metrics}. Our method significantly outperforms other approaches both qualitatively and quantitatively, achieving strong prompt alignment, high aesthetic appeal, and exceptional quality and detail. In addition, our generated humans are fully animatable and support advanced rendering such as subsurface scattering. Our method is also much faster than the other methods, which require from one hour to up to half a day to complete.

\subsection{Comparisons of Text-to-3D Garment Generation}

\begin{table}
    \renewcommand{\arraystretch}{1.1}
    \centering
    \caption{Quantitative comparison and ablation results for text-to-3D garment generation.
    }
    \label{tab:t2g}
    \begin{tabular}{c|l|cccc} 
        \Xhline{2\arrayrulewidth}
        & Method & CLIP & FCLIP & Aesth. & IQ \\ 
        \Xhline{2\arrayrulewidth}
        \multirow{6}{*}{\rotatebox[origin=c]{90}{w/ tex}} 
        & DressCode & 22.42 & 25.64 & 3.53 & 0.596 \\
        & GarmentDreamer & 25.42 & 28.79 & 3.57 & 0.533 \\
        & Ours-\textit{Paint3D} & 26.00 & 31.53 & 3.98 & 0.705 \\
        & Ours-\textit{SyncMVD} & 27.39 & 32.37 & 4.13 & 0.710 \\
        & Ours-\textit{Hunyuan3D-2} & 25.10 & 29.84 & 4.00 & 0.671 \\
        & Ours & \textbf{28.30} & \textbf{33.26} & \textbf{4.34} & \textbf{0.759} \\
        \hline
        \multirow{4}{*}{\rotatebox[origin=c]{90}{w/o tex}} 
        & Garment3DGen & 20.44 & 25.56 & 3.31 & --- \\
        & ChatGarment & 21.39 & 24.66 & 3.50 & --- \\
        & Ours w/o $\mathcal{L}_{coll}$ & 18.38 & 25.84 & 3.66 & --- \\
        & Ours & \textbf{22.01} & \textbf{25.93} & \textbf{3.81} & --- \\
        \hline
        \Xhline{2\arrayrulewidth}
    \end{tabular}
\end{table}

We compare Tailor with state-of-the-art text-to-garment methods to evaluate its superior performance in high-fidelity garment generation. The compared methods include DressCode~\cite{he2024dresscode} and GarmentDreamer~\cite{li2025garmentdreamer}, which support texture generation (w/ tex), as well as Garment3DGen~\cite{sarafianos2024garment3dgen} and ChatGarment~\cite{ChatGarment}, which either do not support or lack available code for texture generation (w/o tex). Since Garment3DGen requires image input rather than text, we supply either real images or images generated by Stable Diffusion 3.5\cite{SD3}, aligned with the corresponding text prompts. We illustrate qualitative comparisons in Fig.~\ref{fig:t2g-1} and Fig.~\ref{fig:t2g-2}. 
Our method excels at generating both accurate, smooth geometry and high-quality, semantically accurate textures that align with text prompts. Note our model's ability to render iconic designs (``Spider-Man bodysuit'') and specific compositional details (``vintage-inspired baseball tee with raglan sleeves'' and ``embroidered initials''). Competing methods often produce noisy or implausible geometries and textures, or fail to accurately follow the text prompt.
For quantitative evaluation, we generate 22 distinct garments from text prompts spanning common garment types, including shirts, pants, skirts, dresses, jackets, and overalls. To assess diversity, we also include fantastical garments such as ``Iron Man armor''. We render the color or shape images of the generated garments from surrounding viewpoints, and calculate the metrics across all views. Quantitative results are summarized in Table~\ref{tab:t2g}. Our method generates accurate, well-defined geometry and natural, high-quality textures with strong prompt adherence, achieving superior performance across all metrics. Please also see  Fig.~\ref{fig:illustration1} 
in the Appendix and the supplementary demo video for more examples, comparisons, and physics-based simulation.

In addition, we compare our texture generation pipeline with state-of-the-art mesh texturing methods, including Paint3D~\cite{zeng2024paint3d}, SyncMVD~\cite{SyncMVD}, and Hunyuan3D-2~\cite{Hunyuan3D-2}. Specifically, we replace our texture module with each of these methods to form the ``Ours-\textit{X}'' baselines, where the geometry generated by our method is used as input and textures are generated under the same text prompts. Qualitative comparisons are illustrated in Fig.~\ref{fig:comparison_texture} and quantitative results are summarized in Table~\ref{tab:t2g}. The comparison shows our clear advantage in generating high-fidelity and semantically accurate textures.

\begin{figure}[ht]
\centering
\includegraphics[width=1\linewidth]{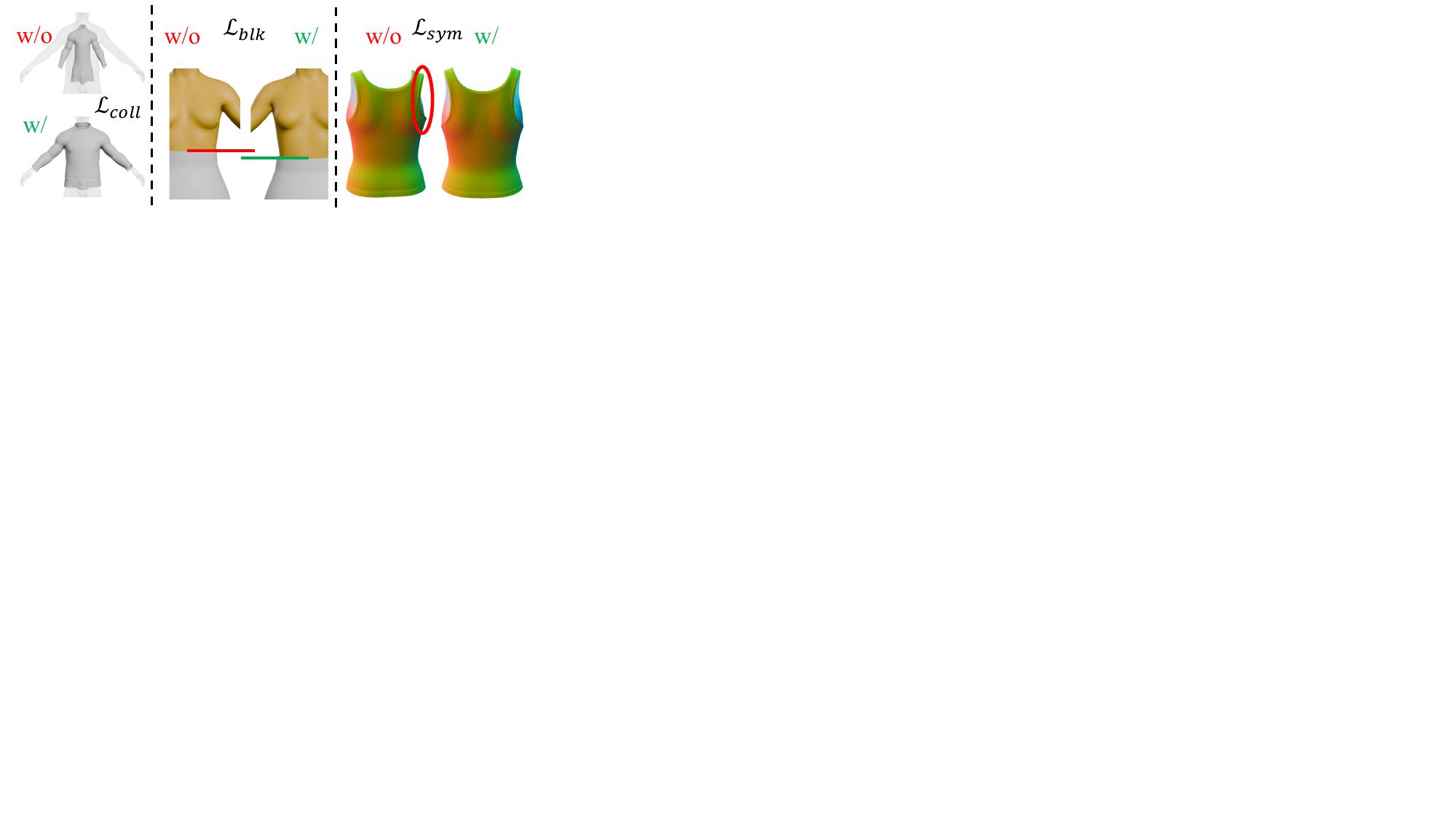}
\caption{Artifacts appear when the geometric constraints are removed.}
\label{fig:ablation_geo}
\end{figure}

\subsection{User Study} \label{ap:user_study}

\begin{table}[htb]
    \centering
    \renewcommand{\arraystretch}{1.1}
    \caption{User study assessing text-to-garment generation quality. Participants rank the results of the three comparison methods.}
    \label{tab:userstudy}
    \begin{tabular}{l|ccc}
    \Xhline{2\arrayrulewidth}
    \multirow{2}{*}{Method} & \multicolumn{3}{c}{Avg. User Ranks} \\
     & GQ & TQ & TA \\
    \Xhline{2\arrayrulewidth}
    GarmentDreamer & 2.72 & 2.34 & 2.06 \\
    DressCode & 2.11 & 2.47 & 2.76 \\
    Ours & \textbf{1.17} & \textbf{1.19} & \textbf{1.18} \\
    \Xhline{2\arrayrulewidth}
    \end{tabular}
\end{table}

We conduct user studies to evaluate the generation quality of both humans and garments. For text-to-human generation, participants evaluate Visual Quality (VQ) and Text Alignment (TA) while viewing multi-view and multi-scale renderings of various body parts alongside the text prompt. For text-to-garment generation, participants evaluate Geometric Quality (GQ), Texture Quality (TQ), and Text Alignment (TA) by viewing 360$^\circ$ RGB and normal map renderings. In both cases, participants rank the competing methods for each metric. Our data includes responses from 25 participants, covering 20 comparisons for humans and 48 for garments. As shown in Table~\ref{tab:t2h_metrics} and Table~\ref{tab:userstudy}, our method receives the highest overall preference.

\subsection{Ablation Study}

\begin{figure}[ht]
\centering
\includegraphics[width=\linewidth]{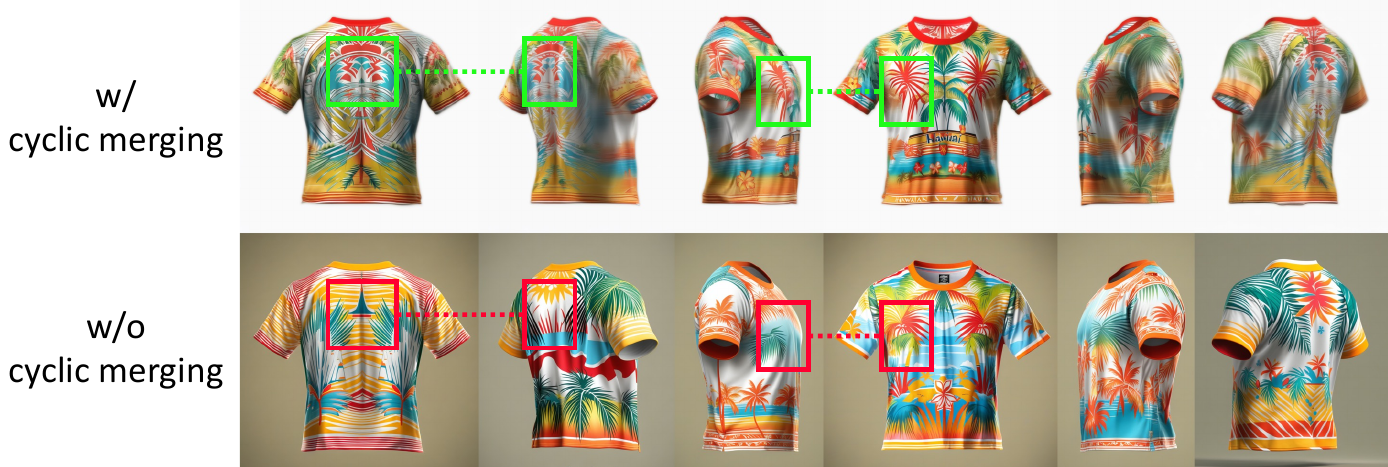}
\caption{Images generated by our multi-view diffusion model. The cyclic merging operation ensures multi-view consistency.}
\label{fig:abl_tex_aggr}
\end{figure}

\begin{figure}[ht]
\centering
\includegraphics[width=0.8\linewidth]{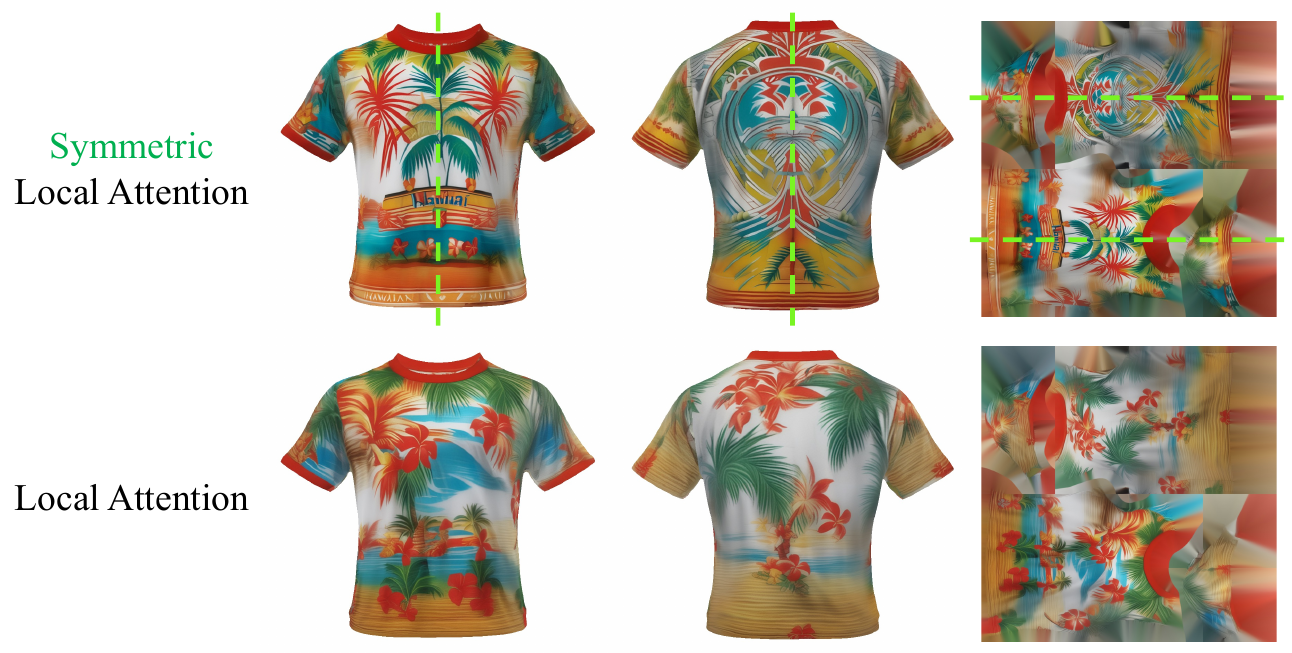}
\caption{Effect of the symmetric local attention.}
\label{fig:abl_tex_sym}
\end{figure}

We conduct ablation studies on key components of both the geometry and texture generation pipelines. We first examine the geometric constraints. As shown in Table~\ref{tab:t2g} and Fig.~\ref{fig:ablation_geo}, removing $\mathcal{L}_{\text{blk}}$ leads to misaligned clothing and a decline in all evaluation metrics. This loss not only encourages alignment with the body and prevents arment-body penetrations, but also aids garment generation by providing the human body as an indirect prior. The two auxiliary losses, $\mathcal L_{coll}$ and $\mathcal L_{sym}$, improves symmetry and prevent garments from excessive growth when applied. We also illustrate the effect of the cyclic merging technique and symmetric local attention in Fig.~\ref{fig:abl_tex_aggr} and Fig.~\ref{fig:abl_tex_sym}, respectively.
The cyclic merging technique ensures multiview consistency, while the attention mechanism ensures symmetric texture in a flexible manner.

\section{Discussion}

\textbf{System Extensibility.} Tailor acts as a training-free framework for the joint generation of CG-ready humans and garments. Designed for customizability and extensibility, the framework allows users to effortlessly expand current instruction prompts and template libraries to accommodate diverse requirements. Its modular architecture further supports broad implementation flexibility, enabling users to reconstruct the system using alternative human generators (e.g., MetaHuman), LLMs, or image diffusion models to suit evolving production needs.

\textbf{Limitations.} Despite these promising results, our approach exhibits several limitations. First, while our method excels at geometric adaptation and stylistic refinement, it may yield suboptimal results for requests requiring significant topological changes or structural additions and deletions relative to the template. However, thanks to the extensibility of our training-free framework, we can mitigate this by expanding the template coverage, including incorporating the system's own successful generations into the library. Second, our method is presently limited to single-piece clothing and does not support nested garments. Additionally, our method is not suitable for generating non-manifold structures, such as clothing pockets. These challenges offer compelling avenues for future research, and we plan to address them in our subsequent work.

\section{Conclusion}

In this paper, we present Tailor, an integrated framework for generating high-fidelity 3D human avatars with disentangled clothing from textual descriptions. The generated characters can be exported to standard file formats for compatibility with third-party software and can be readily integrated into CG software for animation and simulation, thus enabling novel applications in fields such as virtual try-on and digital animation. Experiments demonstrate that our pipeline outperforms existing text-to-3D generation methods in texture and geometry quality and alignment with text descriptions.


\bibliographystyle{IEEEtran}
\bibliography{main}

\vfill
\clearpage
\appendix
\counterwithin{figure}{section}  
\counterwithin{table}{section}
\setcounter{figure}{0}
\setcounter{table}{0}
\setcounter{page}{1}
\section*{Additional Qualitative Comparisons} \label{appendix:additional_comparisons}

\begin{figure*}[b]
\centering
\includegraphics[width=\linewidth]{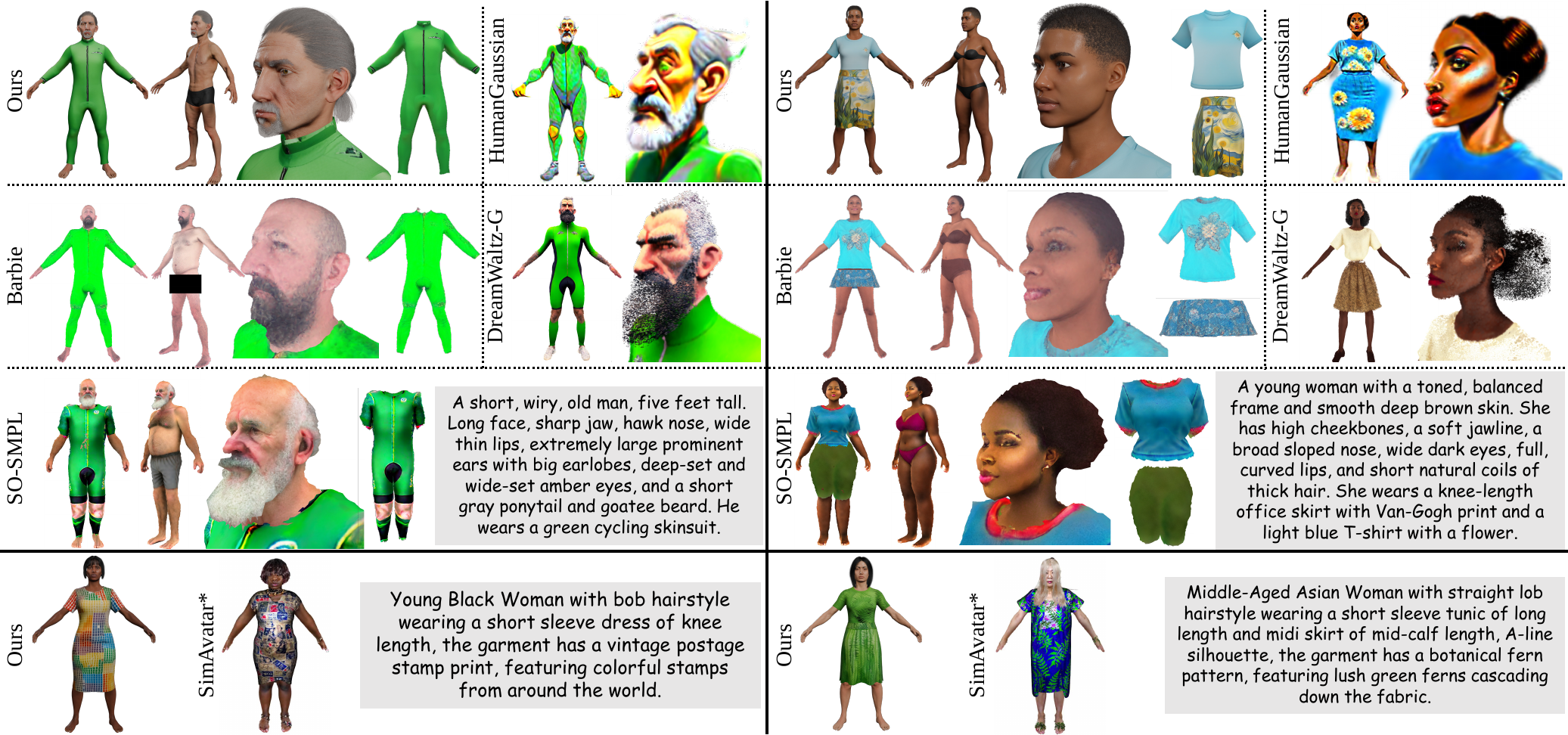}
\caption{Additional qualitative comparison with state-of-the-art text-to-3D human generation methods.} \label{fig:t2h-2}
\end{figure*}

\begin{figure*}[htbp]
\centering
\includegraphics[width=\linewidth]{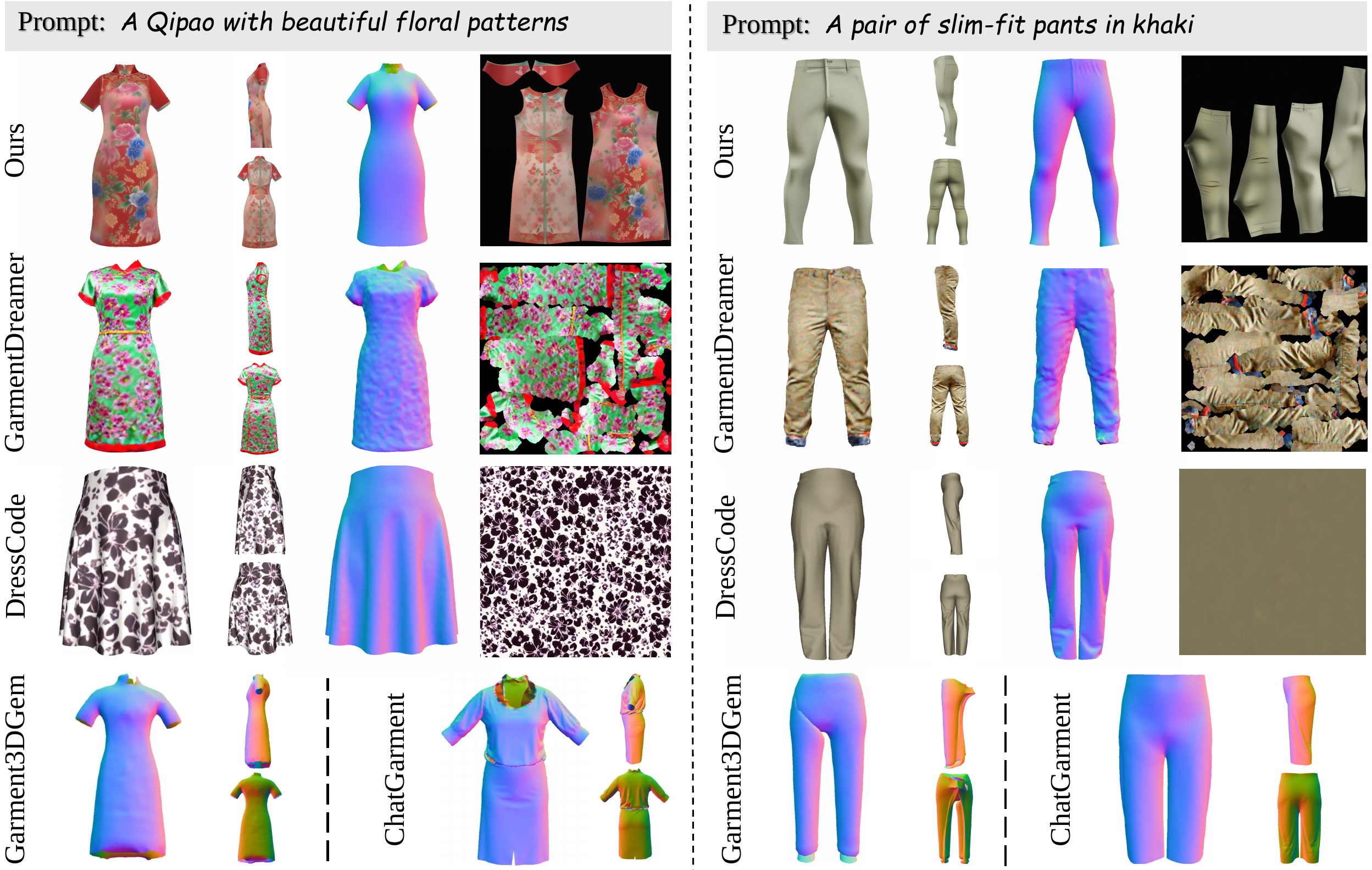}
\caption{Additional qualitative comparison with state-of-the-art text-to-3D garment generation methods.} \label{fig:t2g-2}
\end{figure*}

We show additional qualitative comparisons of text-to-human and text-to-garment generation, illustrated in Fig.~\ref{fig:t2h-2} and Fig.~\ref{fig:t2g-2}.

\section*{Additional Results} \label{appendix:additional_results}

\begin{figure*}[htbp]
\centering
\includegraphics[width=\linewidth]{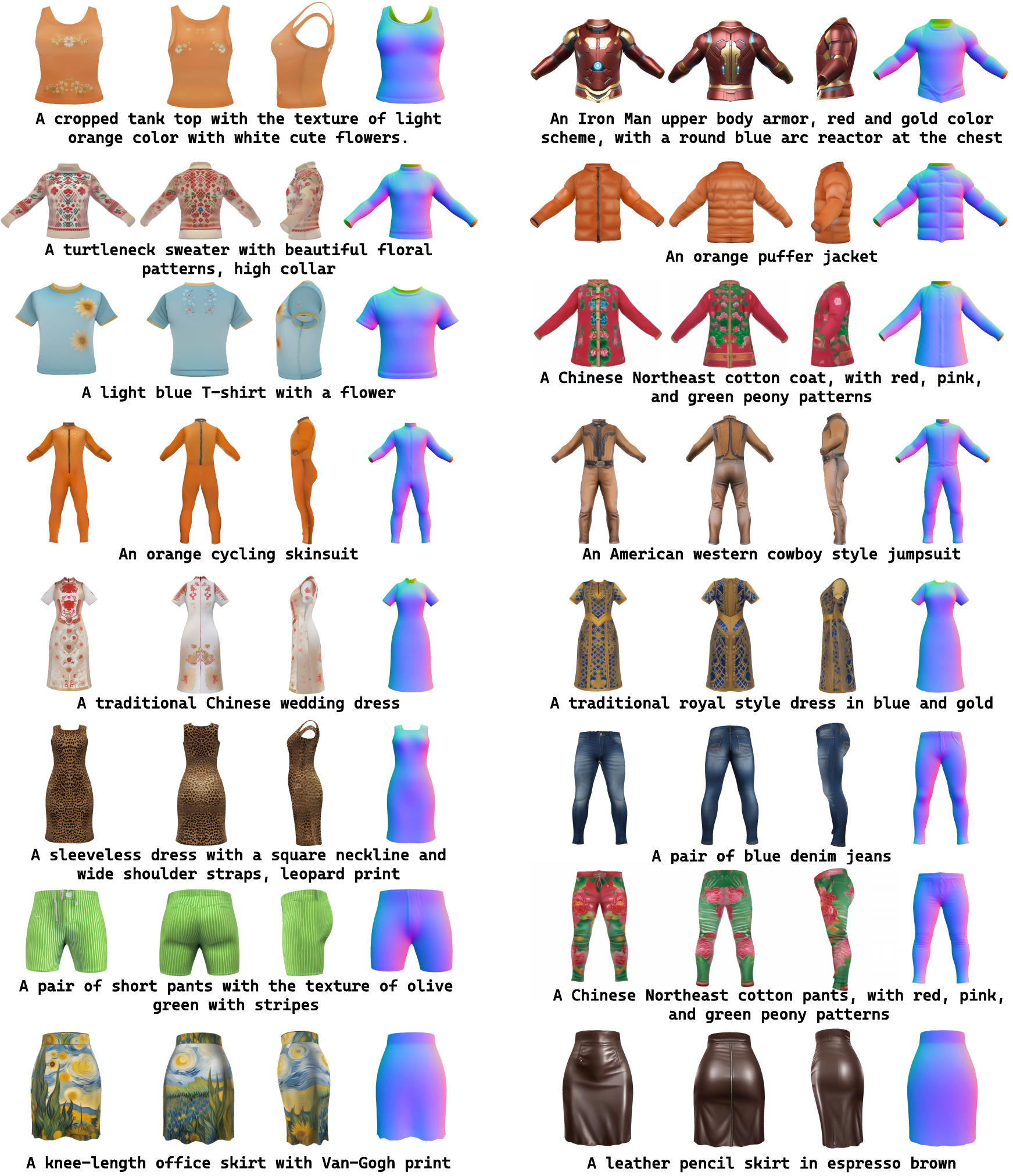}
\caption{Additional results of generated garments from our method.} \label{fig:illustration1}
\end{figure*}

We present the RGB and normal images of additional garments generated by our method in Fig.~\ref{fig:illustration1}. The garments produced by our approach exhibit high quality and diverse geometry and texture.

\section*{Prompts for the LLM Agent} \label{appendix:llm_prompts}
We carefully designed eight instruction prompts, employing in-context learning to enable the Large Language Model to perform the tasks of Prompt Decomposition, Body Parameterization, and Garment Template Matching, as described in Sec.~\ref{subsec:hum_gen_llm}. 

\begin{figure*}[ht]
\centering
\includegraphics[width=1\linewidth]{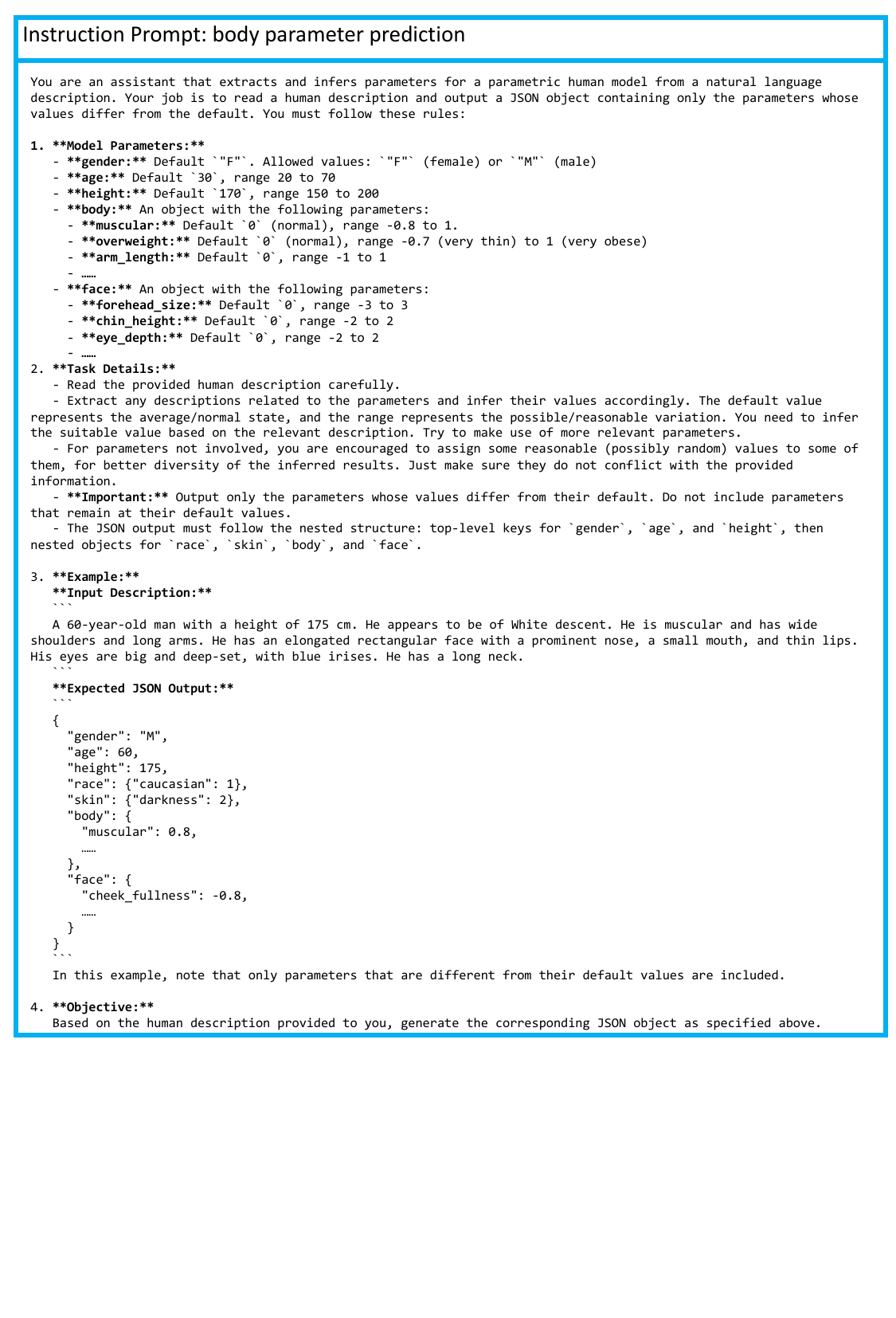}
\caption{The instruction prompt for predicting body parameters from body description.} \label{fig:prompt2}
\end{figure*}

\begin{figure*}[ht]
\centering
\includegraphics[width=1\linewidth]{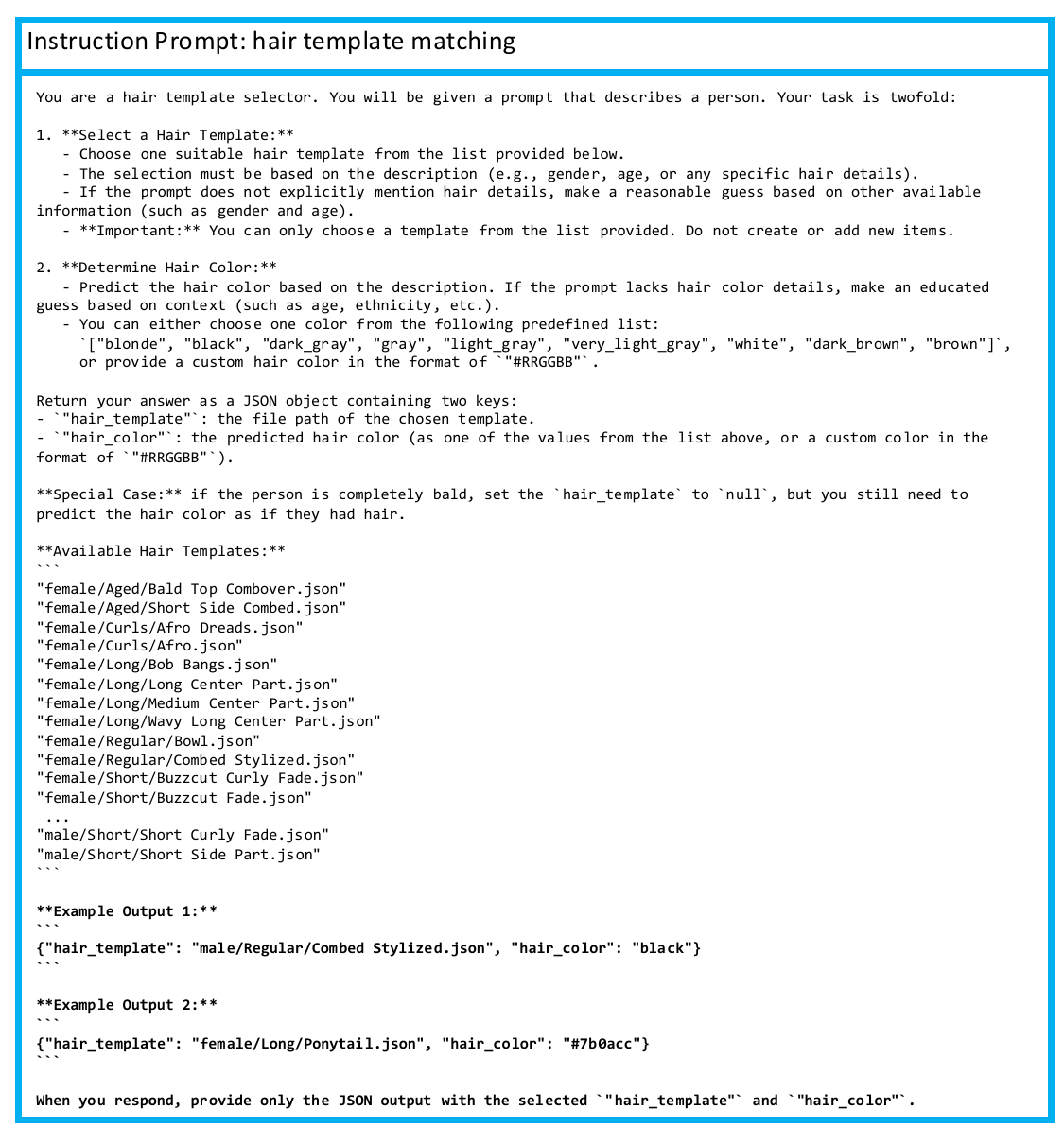}
\caption{The instruction prompt for predicting hair template and color from body description.} \label{fig:prompt_hair}
\end{figure*}

The core content of several LLM instruction prompts is presented below. Figure~\ref{fig:prompt1} illustrates the prompt for decomposing the user-provided description of a human into separate body and garment sub-prompts. The resulting sub-prompts are then processed independently in subsequent stages. Figure~\ref{fig:prompt2} shows the instruction prompt that directs the LLM to translate the body description into HumGen3D's body parameters. This prompt defines the parameter names, valid ranges, and corresponding descriptions, and provides detailed instructions and an example to ensure accurate parameter prediction. In addition, Figure~\ref{fig:prompt_hair} shows how the hair template and color are determined.

\begin{figure*}[ht]
\centering
\includegraphics[width=1\linewidth]{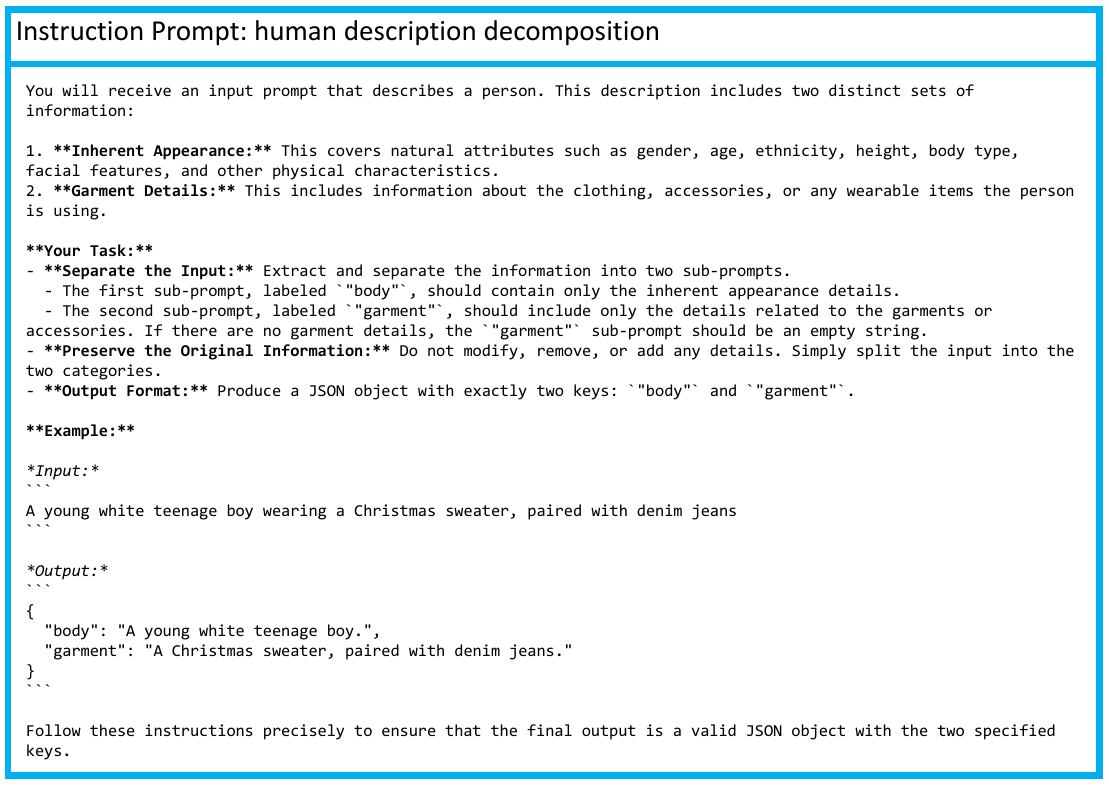}
\caption{The instruction prompt for decomposing the overall human description into body and garment sub-prompts.} \label{fig:prompt1}
\end{figure*}

\begin{figure*}[ht]
\centering
\includegraphics[width=1\linewidth]{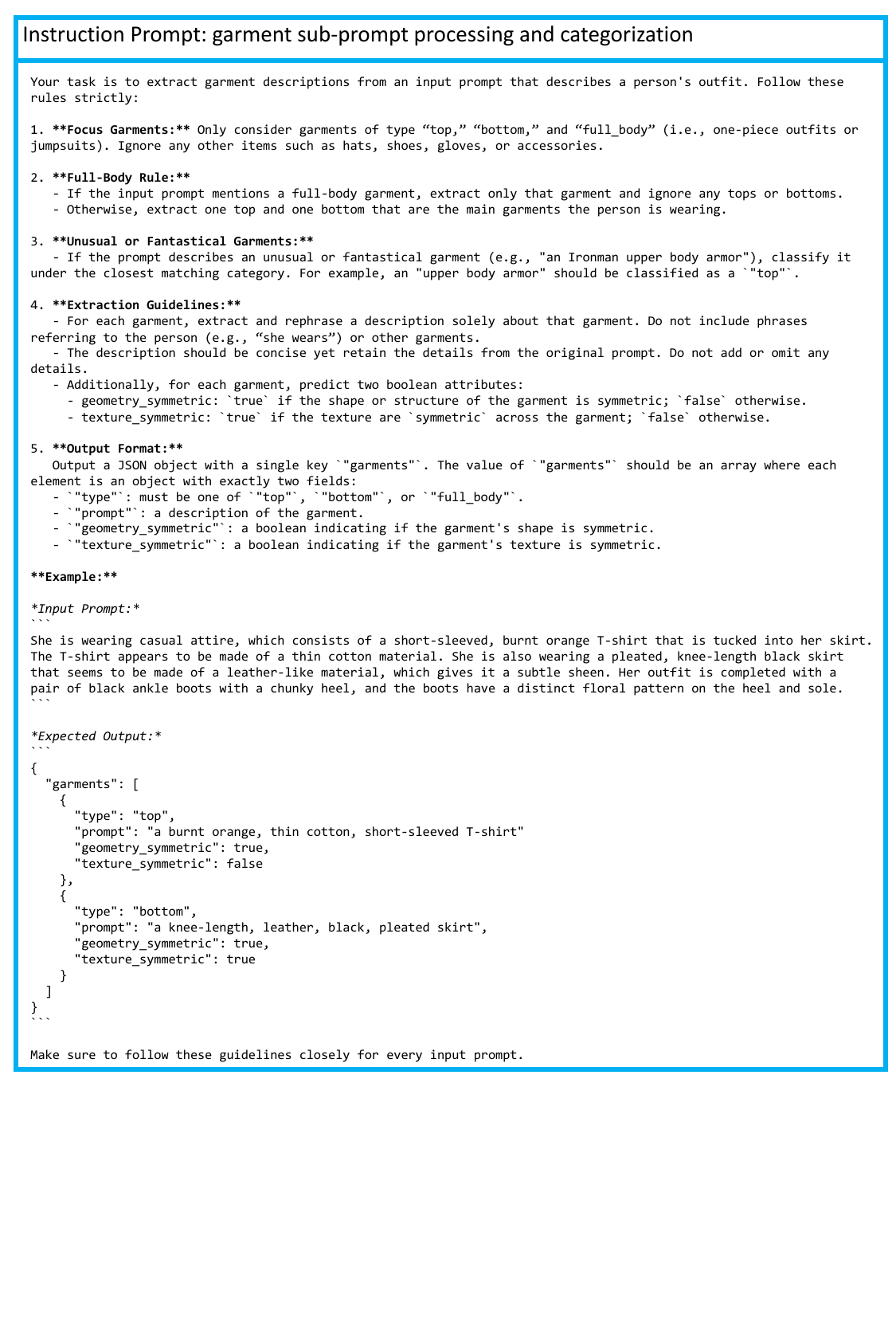}
\caption{The instruction prompt for garment sub-prompt processing and categorization.} \label{fig:prompt4}
\end{figure*}

\begin{figure*}[ht]
\centering
\includegraphics[width=1\linewidth]{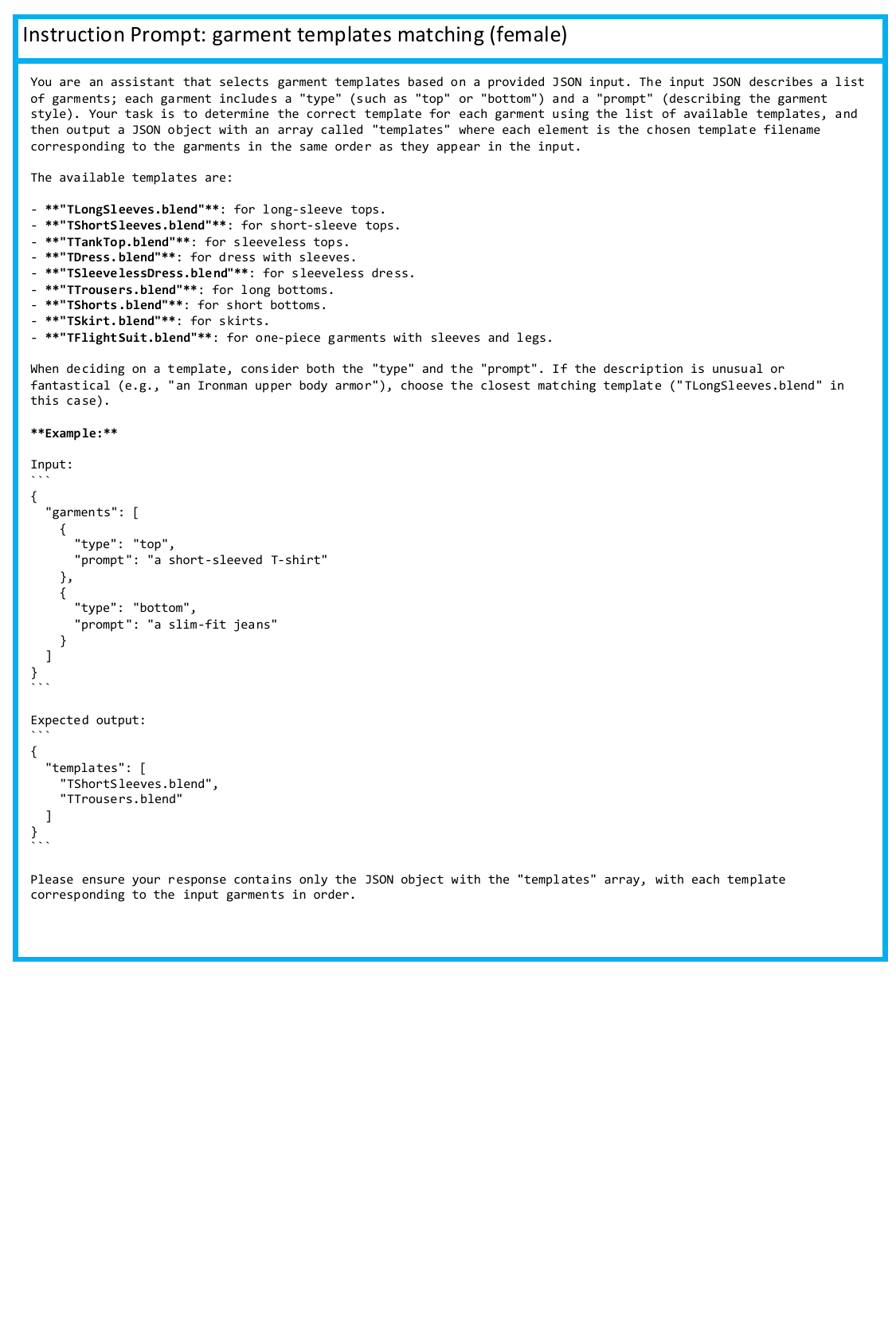}
\caption{The instruction prompt for garments template matching.} \label{fig:prompt5}
\end{figure*}

The garment sub-prompts undergo further processing to provide garment templates and descriptions for garment geometry generation (Sec.~\ref{subsec:garment_geo_gen}) and texture generation (Sec.~\ref{subsec:texture}). The LLM first selects either a full-body garment or an outfit consisting of a top and a bottom garment, then extracts the corresponding descriptions and assigns category labels based on the garment type (see Fig.~\ref{fig:prompt4}). Additionally, the LLM extracts pure geometric descriptions of the garment. It then selects the closest match garment template for each extracted garment. Fig.~\ref{fig:prompt5} shows the instruction prompt for matching garments template for females. This prompt lists the available templates for matching and provides an example. The corresponding prompt for matching garment templates for male is similar.

The LLM also processes hair and facial hair information from the body description and selects the closest matching templates in a manner similar to garment template matching.

\end{document}